\documentclass[preprint,12pt]{elsarticle}

\usepackage{lineno,hyperref}
\usepackage{amssymb,amsmath, amsthm}
\usepackage{multirow}
\usepackage{algorithm}
\usepackage{algpseudocode}
\usepackage{subcaption}

\journal{xxxx}

\begin{document}
	
	\begin{frontmatter}
		
		
		
		\title{RANG: A Residual-based Adaptive Node Generation Method for Physics-Informed Neural Networks}
		
		
		\author[inst1]{Wei Peng}
		\ead{weipeng0098@126.com}
		\affiliation[inst1]{organization={Defense Innovation Institute, Chinese Academy of Military Science},
			city={Beijing},
			postcode={100071}, 
			country={China}}
		\author[inst1]{Weien Zhou}\ead{weienzhou@nudt.edu.cn}
		\author[inst1]{Xiaoya Zhang}
		\author[inst1]{Wen Yao\corref{cor}}
		\ead{wendy0782@126.com}
		\cortext[cor]{Corresponding Author}
		
		\author[inst2]{Zheliang Liu}
		
		\affiliation[inst2]{organization={Human Resource Center},
			city={Beijing},
			postcode={100031}, 
			country={China}}
		
		\begin{abstract}
			Learning solutions of partial differential equations (PDEs) with Physics-Informed Neural Networks (PINNs) is an attractive alternative approach to traditional solvers due to its flexibility and ease of incorporating observed data. 
			Despite the success of PINNs in accurately solving a wide variety of PDEs, the method still requires improvements in terms of computational efficiency.
			One possible improvement idea is to optimize the generation of training point sets.
			Residual-based adaptive sampling and quasi-uniform sampling approaches have been each applied to improve the training effects of PINNs, respectively.
			To benefit from both methods, 
			we propose the Residual-based Adaptive Node Generation (RANG) approach for efficient training of PINNs, which is based on a variable density nodal distribution method for RBF-FD.
			The method is also enhanced by a memory mechanism to further improve training stability.
			We conduct experiments on three linear PDEs and three nonlinear PDEs with various node generation methods, through which the accuracy and efficiency of the proposed method compared to the predominant uniform sampling approach is verified numerically. 
			
		\end{abstract}
		
		
		
		
	\end{frontmatter}
	
	
	\section{Introduction}
	When solving PDEs, generating a high-quality mesh can be computationally expensive. Therefore, to avoid generating any mesh, many meshless methods have been developed to solve partial differential equations numerically in recent years. Meshless methods usually generate the computational nodes to fit the boundary and meet the spatially variable resolution requirement, thus avoiding generating any mesh. Due to the high freedom of geometry, meshless methods have great potential in dealing with data-driven related or inverse PDE problems even with complex domains, especially in recent methods incorporating deep neural networks.
	
	Deep Neural Networks (DNN) have achieved remarkable performance in many scientific computing tasks, including recent studies using DNN models to solve PDEs numerically. This paper focuses on one of these methods, the Physics-Informed Neural Network (PINN) \cite{raissiPhysicsinformedNeuralNetworks2019}, where physical conditions are imposed on a feedforward neural network. Though PINN usually appears as a deep learning method,  it is also a strong-form meshless PDE solver.
	
	PINN has arisen as a promising framework for combining observed data and physical laws for various applications in science and engineering. 
	Raissi et al. proposed PINN to identify unknown parameters in partial differential equations (PDEs), such as the Korteweg-de Vries equation \cite{raissiPhysicsinformedNeuralNetworks2019}, and used the Navier-Stokes equations to reveal hidden fluid dynamics \cite{raissiHiddenFluidMechanics2020}. 
	Jin et al. \cite{jinNSFnetsNavierStokesFlow2021} employed PINN to model incompressible flow from laminar to turbulent. 
	Sun et al. \cite{sunSurrogateModelingFluid2020a} proposed PINN-based surrogates of fluids without relying on any external data. 
	Haghighat et al. \cite{haghighatPhysicsinformedDeepLearning2021} used PINN for inversion and surrogate modeling in the field of solid mechanics. 
	Cai et al. \cite{caiPhysicsInformedNeuralNetworks2021} investigated solutions to various prototype heat transfer problems. 
	He et al. \cite{hePhysicsinformedDeepLearning2021} studied the problem of direct and inverse heat conduction in materials.
	Lu et al. \cite{luPhysicsInformedNeuralNetworks2021} used PINN in reverse design to solve the holographic problem in optics and the fluid problem of Stokes flow. 
	Schiassi et al. \cite{schiassiPhysicsInformedNeuralNetworks2022} used PINN to solve the optimal planar orbital transfer problem. 
	PINN is also used to generate grid meshes for traditional methods \cite{chenMGNetNovelDifferential2022a}.

	Since a DNN is a nonlinear function containing many parameters, PINN discretizes a PDE by modeling it as an unconstrained nonlinear least-squares optimization problem. However, PINN often faces serious computational difficulties and fails to solve the PDE due to the high nonlinearity. Therefore, a series of extensions to the vanilla PINN has been proposed to improve the accuracy and robustness of handling increasingly challenging problems. Such extensions include, but are not limited to, tuning of loss weights \cite{wangUnderstandingMitigatingGradient2021, mcclennySelfAdaptivePhysicsInformedNeural2020, xiang2021selfadaptive, liuDualDimerMethodTraining2021}, novel network architectures and activation functions \cite{ramabathiranSPINNSparsePhysicsbased2021a, wangRespectingCausalityAll2022, yuGradientenhancedPhysicsinformedNeural2022, cuiEfficientNeuralNetwork2022}, domain decomposition methods \cite{mengPPINNPararealPhysicsinformed2020}, and mini-batch collocation generation algorithms \cite{dasStateoftheArtReviewDesign2022, nabianEfficientTrainingPhysics2021a, luDeepXDEDeepLearning2021, zhaoSolvingAllenCahnCahnHilliard2021}.
	
	From another perspective, PINN is a meshless method with DNNs replacing RBFs, which is a direct extension of the Kansa method \cite{kansaMultiquadricsScatteredData1990}. For traditional meshless approaches, especially RBF-FD methods, the choice of collocation points requires constraints on node sets to ensure solution accuracy and stability \cite{fornbergSolvingPDEsRadial2015}. This paper will draw on extensive research on node generation approaches in the traditional meshless field. Current methods of node generation can be categorized broadly into iterative methods, sphere packing methods, and advancing front methods. Iterative methods begin with initializing a node set and updating their positions \cite{zamoloTwoAlgorithmsFast2018, vlasiukFastHighdimensionalNode2018}, which are computationally expensive and require an initial distribution. Sphere packing methods are often constrained to a constant radius, which leads to constant node density. One of the most usual methods is Poisson disk sampling \cite{bridsonFastPoissonDisk2007}. Advancing front methods are computationally more efficient and relatively simple to implement. The proposed method in this paper is based on the original advancing front type method presented by Fornberg and Flyer \cite{fornbergFastGeneration2D2015}. Other similar works include \cite{vandersandeFastVariableDensity2021, duhFastVariableDensity2021, slakGenerationNodeDistributions2019a,liBitingAdvancingFront2000, lhnerProgressGridGeneration1996}. The meshless method RBF-FD can also be combined with adaptive node generation strategies. For example, Li et al. \cite{liAdaptiveRBFFDMethod2017} and Slack et al. \cite{slakAdaptiveRadialBasis2019} proposed locally adaptive sampling methods based on the estimation of PDE residuals.
	
	\textbf{Related work}. There are relatively few studies on the generation of PINN collocation points. Lu et al. \cite{luDeepXDEDeepLearning2021} proposed the RAR method, which gradually adds and retains the collocation points with more significant residuals in the collocation point set.  \cite{nabianEfficientTrainingPhysics2021a} employed the mode of subdomain training to strengthen the training of regions with large residuals. Another subdomain adaptive sampling method is proposed by Zeng et al.  \cite{zengAdaptiveDeepNeural2022}. \cite{zhaoSolvingAllenCahnCahnHilliard2021} found it challenging for the vanilla PINN to solve the Allen-Cahn and Hilliard equations and improved results after adopting a similar RAR strategy. 
	Tang et al. \cite{tang2021} introduced an additional neural network acting as an error indicator to guide the refinement, which is aimed at avoiding the large variance in uniform random sampling. Hanna et al. \cite{hannaResidualbasedAdaptivityTwophase2022} estimated probalistic density functions to adaptively generate addtitional nodes based on the PDE residuals.
	From the experimental design perspective, Das and Tesfamariam \cite{dasStateoftheArtReviewDesign2022} discussed the effects of various sampling methods and compared several node generation methods for PINN. In their experimental results, the quasi-uniform method, Hammersley sampling, seems very simple but has a significant advantage over other methods. 
	
	\textbf{Motivation}. These studies inspire us that the combination of quasi-uniform sampling and residual-based adaptive sampling may further improve PINN training.
	However, the adaptability and quasi-uniformity of a sampling method seem to be two contradictory goals, and this paper is trying to reconcile the contradiction by introducing variable density node generation methods with local regularity. Based on the rapid node generation technology in two-dimensional space proposed by \cite{fornbergFastGeneration2D2015}, it is possible to generate locally quasi-uniform nodes with variable density based on the PDE residuals adaptively. The Residual-based Adaptive Node Generation method (RANG) is then proposed via combining the residual of PINN with the method that can dynamically change the sampling radius in a computational domain. Therefore, on the one hand, we expect to take advantage of the quasi-uniform sampling method that may bring to PINN and, on the other hand, combines the advantages brought by residual-based adaptive refinement methods.
	
	The structure of this paper is organized as follows. Section 2 introduces PINNs and two basic sampling techniques. Section 3 proposes RANG for PINNs. Section 4 conducts six numerical experiments, including three linear and three nonlinear PDEs. The final section discusses the experimental results, points out the limitations of this paper and possible improvements, and concludes the paper.
	
	\section{Background}
	This section will introduce the necessary elements for solving PINN with RANG. First, the vanilla PINN is briefly introduced. Then two sampling methods that generate nodes in 2D rectangles are introduced, where the Hammersley sampling method is introduced for reference, and an advancing front sampling method is introduced to be the basis of RANG.
	\subsection{Physics Informed Neural Networks}
	PINN employs a feedforward neural networks $\hat u$ to approximate the solution $u$ for the following generic differential equation:
	\begin{align*}
		\mathcal{N}(t,x,u)=0, &\quad t\in[0,T],x\in D,\\
		\mathcal{I}(x,u)=0, &\quad x\in D,\\
		\mathcal{B}(t,x,u)=0, &\quad t\in[0,T],x\in\partial D,
	\end{align*}
	where $\mathcal{N}$ is a general differential operator that may consist of derivatives,  linear terms, and nonlinear terms. $x$ usually denotes a spatial vector, while $t$ is a scalar about time. The operators $\mathcal{I}$ and $\mathcal{B}$ denote the initial and boundary value conditions, respectively, which may also consist of differential, linear, and nonlinear terms.
	
	Following the work of Raissi et al. \cite{raissiPhysicsinformedNeuralNetworks2019}, we consider $\hat u_\theta$ to be a classic fully-connected neural network which is parameterized by $\theta$. Define the residual networks, which share the same network parameters $\theta$ and satisfy
	\begin{align*}
		r_{pde}(t,x;\theta)&:=\mathcal{N}(t,x,\hat u(t,x;\theta)),\\
		r_{0}(t,x;\theta)&:=\mathcal{I}(x,\hat u(t,x;\theta)),\\
		r_{b}(t,x;\theta)&:=\mathcal{B}(t,x,\hat u(t,x;\theta)),
	\end{align*}
	where all partial derivatives can be computed by automatic differentiation methods. The shared network parameters $\theta$ are trained by minimizing loss terms that penalizes the residuals for not equaling zero:
	\begin{align}
		L_{pde}(\theta)&=\frac{1}{\#\mathbf{N}_{pde}}\sum_{(x,t)\in \mathbf{N}_{pde}} |r_{pde}(t,x,\theta)|^2,\label{eq:pde_l}\\
		L_{0}(\theta)&=\frac{1}{\#\mathbf{N}_{0}}\sum_{(x,t)\in \mathbf{N}_{0}} |r_{0}(t,x,\theta)|^2,\nonumber\\
		L_{b}(\theta)&=\frac{1}{\#\mathbf{N}_{b}}\sum_{(x,t)\in \mathbf{N}_{b}} |r_{b}(t,x,\theta)|^2, \nonumber
	\end{align}
	where $\mathbf{N}_{pde}$ is the collocation point set for the governing PDE, $\mathbf{N}_0$ represents the initial condition point set, and $\mathbf{N}_b$ represents a boundary point set. The notation $\#$ denotes the number of elements in a set. The total loss is defined by summing up the weighted loss terms:
	\begin{align}\label{eq:total_loss}
		L(\theta)=w_{pde}L_{pde}(\theta)+w_{0}L_{0}(\theta)+w_{b}L_b(\theta),
	\end{align}
	where $w_{pde}$, $w_{0}$, and $w_b$ are positive weights for each loss term. In the case of an inverse problem, the loss term for observed data $L_D(\theta)$ is appended to Eq.\eqref{eq:total_loss}. Finally, the solution $\hat u_\theta$ can be approximated by minimizing the total loss $L(\theta)$:
	\begin{align}\label{eq:minimize}
		\theta^\ast=\arg\min_\theta L(\theta).
	\end{align}
	
	In the vanilla PINN implementation, the PDE loss collocation node set $\mathbf{N}_{pde}$ is often generated by one-time sampling, and then the set is used during the whole training stage. Subsequent studies have pointed out that changing the point set $\mathbf{N}_{pde}$ during training may improve results. 
	The following numerical experiments also show that resampling the collocation node set $\mathbf{N}_{pde}$ after each iteration of a certain number of steps usually leads to a better result than the one-time sampling strategy.

	\subsection{Hammersley Sampling}
	The selection of $\mathbf{N_{pde}}$ seems to play an important role in the training of PINN. 
	An i.i.d. uniform random sampling strategy is usually employed to generate the collocation point set to solve Eq.\eqref{eq:minimize}. Existing software packages and implementations provided by previous studies also include the sampling strategies of quasi-uniform low-discrepancy sampling. The original implementation of PINN \cite{raissiPhysicsinformedNeuralNetworks2019} used Latin Hypercubic Sampling (LHS). IDRLnet \cite{Peng2021IDRLnetAP} provides jittered grid sampling. The authors of \cite{dasStateoftheArtReviewDesign2022} explored the construction of point sets from the perspective of experimental design, compared various methods, and found that Hammersley sampling works well. In the PINN library SimNet \cite{hennighNVIDIASimNetAIAccelerated2021}, Halton pseudo-random sequence is provided as an alternative algorithm for generating low-discrepancy points. Since Hammersley sampling is one of the motivations of this paper and for the subsequent experiments, we briefly introduce Hammersley sampling.
	
	Hammersley sampling is a quasi-uniform sampling method with lower discrepancy than i.i.d. uniform random sampling. Given the number of samples $N$ and an arbitrary prime $p$, we can use Hammersley sampling to generate a point set in a 2D square $[0,1]\times[0,1]$:
	\begin{align*}
		\mathbf{N}=\left\{\left(\frac{k}{N}, \Phi_p(k)\right) | k=1,2,\cdots,N\right\}
	\end{align*}
	where $\Phi_p$ is the Van de Corput sequence which maps an integer
	\begin{align*}
		a_0+a_1p+a_2p^2+\cdots+a_rp^r
	\end{align*}
	to a decimal
	\begin{align*}
		a_0p^{-1}+a_1p^{-2}+a_2p^{-3}+\cdots+a_rp^{-r-1}.
	\end{align*}
	Sampling points in an arbitrary rectangle can be achieved by scaling the square.
	
	This simple approach seems quite effective for PINN training. We conjecture that the low-discrepancy property or local-regularity makes the points evenly distributed on the computational domain, which increases the quality of the collocation point set. 
	On the other hand, the residual-based sampling method locally refines points on the high residual region and can also improve the efficiency of PINN training. 
	However, it is difficult for  Hammersley sampling to vary the spatial density and thus hard to implement residual-based refinement directly. 
	Therefore, in the following subsection, we introduce a variable density high-quality node generation method used in meshless methods to combine these two aspects of potential advantages.
	
	\subsection{Node Generation by Fornberg and Flyer}
	
	Although Hammersley sampling seems effective in PINN, the quality of the generated collocation points is not suitable for traditional meshless solvers such as RBF-FD. Instead, Fornberg and Flyer proposed a node positioning algorithm in 2015 \cite{fornbergFastGeneration2D2015}. The foundamental form of the algorithm constructs 
	a point set in 2D rectangles and is described by Algorithm \ref{alg:ff}. 
	In the following discussion, we use the first letters of the two authors’ surnames (FF) to refer to the algorithm for simplicity.
	
	FF is an advancing node generation method with variable density. Although there is no guarantee for the lower bound of the minimum distance of generated nodes, it still has substantially local regularity.
	
	\begin{figure}[ht]
		\begin{subfigure}{.3\textwidth}
			\centering
			\includegraphics[width=1.\linewidth]{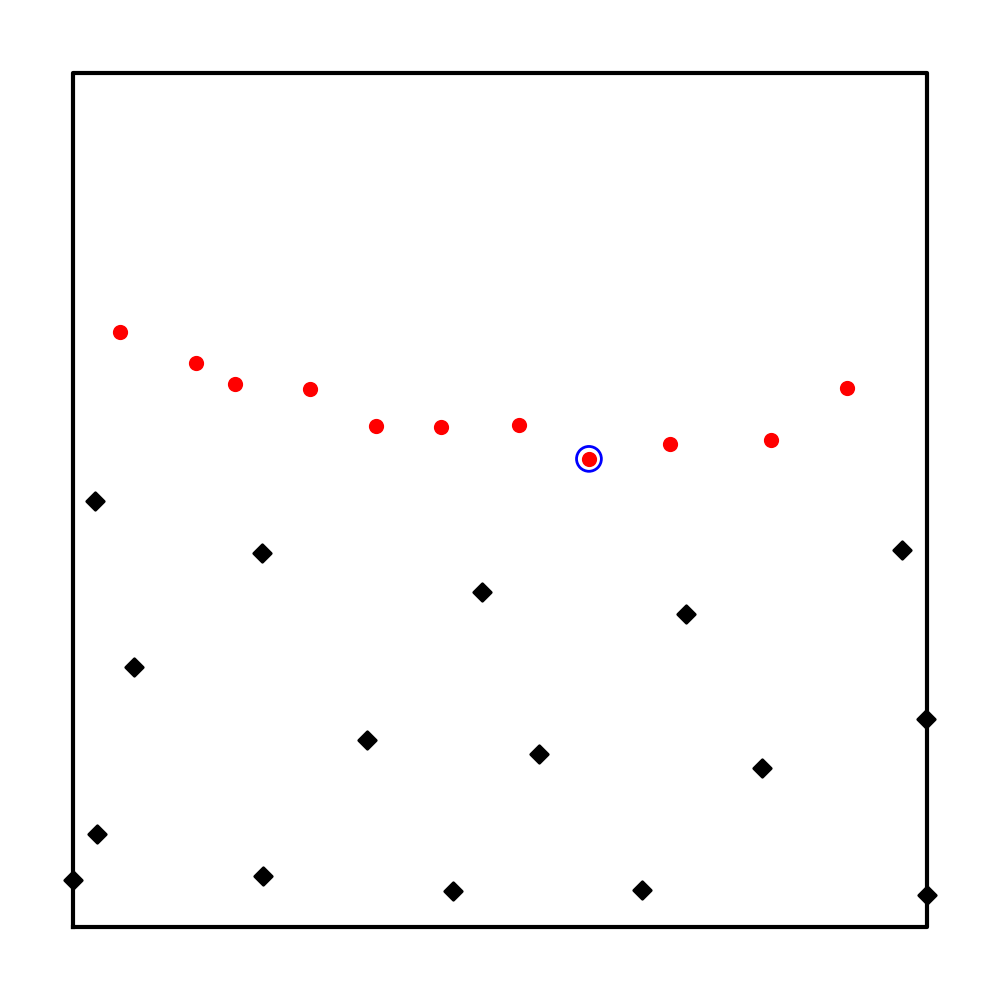}  
			\caption{}
			\label{fig:ff1}
		\end{subfigure}
		\begin{subfigure}{.3\textwidth}
			\centering
			\includegraphics[width=1.\linewidth]{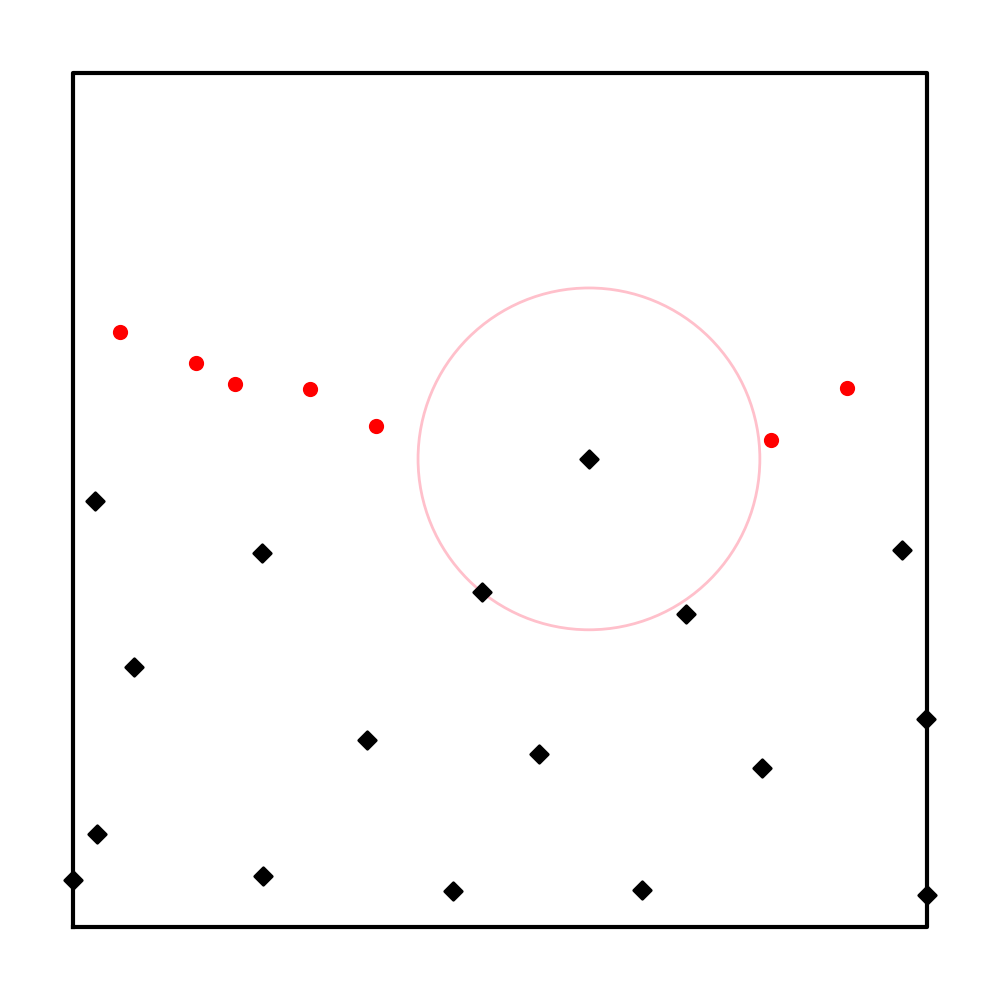}  
			\caption{}
			\label{fig:ff2}
		\end{subfigure}
		\begin{subfigure}{.3\textwidth}
			\centering
			\includegraphics[width=1.\linewidth]{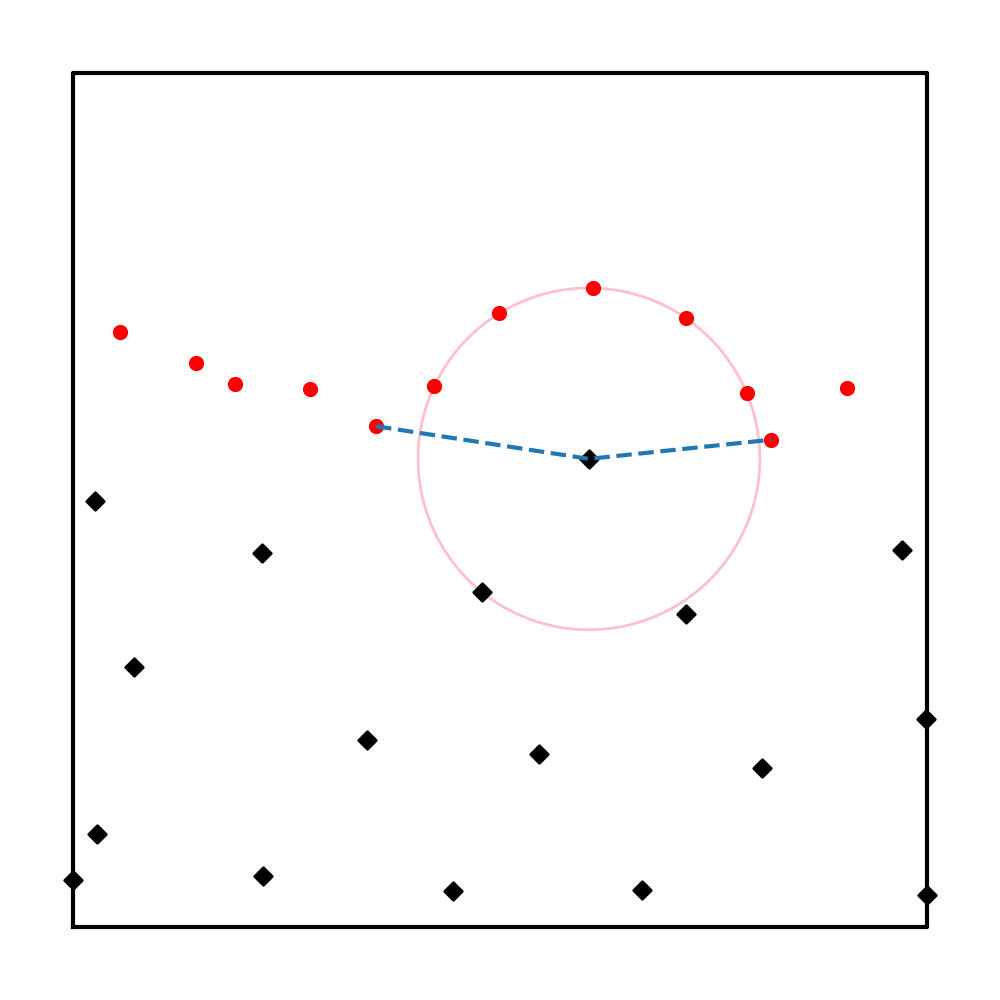}  
			\caption{}
			\label{fig:ff3}
		\end{subfigure}
		\caption{Illustration of Algorithm \ref{alg:ff}. The potential nodes are marked as solid red circles. The generated nodes are marked black diamonds. (a) Identify the lowest potential node in the $y$-axis direction. (b) Delete potential nodes within distance $r$. (c) Generate new potential nodes.}
		\label{fig:ff}
	\end{figure}
	
	The core part of the FF method is sampling points in a 2D rectangle. Two sets of nodes are maintained. One set is the potential node set, and the other is the selected node set. An overview of the method is shown in Figure \ref{fig:ff}. The main steps are listed as follows.
	\begin{enumerate}
		\item Randomly initialize some nodes at the bottom of the rectangular as the potential point set $A$, and initialize the selected node set $B$ to be an empty set.
		\item Find the lowest node in the $y$-axis direction from the potential node set $A$, and add the point to $B$. (Figure \ref{fig:ff1})
		\item Remove all the potential points from $A$ with a distance less than $h(\hat x,\hat y)$ to the selected potential point. The radius can vary across spatial locations. (Figure \ref{fig:ff2})
		\item The two neighboring points on the left and right sides of the node determine an arc with a radius of $h(\hat x,\hat y)$. Generate $5$ new equispaced nodes across the arc and add them to $A$. (Figure \ref{fig:ff3})
		\item Repeat the above the 2-4 steps until the selected potential node exceeds the upper bound.
	\end{enumerate}
	The pseudo-code is given in Algorithm \ref{alg:ff}.
	\begin{algorithm}
		\caption{Node positioning algorithm by Fornberg and Flyer.}\label{alg:ff}
		\begin{algorithmic}[1]
			\Require Box $[x_{min}, x_{max}]\times[y_{min},y_{max}]$,
			\Statex Radius function $h: [x_{min}, x_{max}]\times[y_{min}, y_{max}]\rightarrow (0,\infty)$
			\Statex \textbf{Output} A set of points in the box distributed according to $h$.
			\Function{FF}{$x_{min}, x_{max}, y_{min}, y_{max}, h$}
			
			\State $B \gets $ an empty set of points;
			\State $A\gets $ points spaced according to $h$ from $x_{min}$ to $x_{max}$ at $y$ coordinate $y_{min}$ with small positive perturbation for each point;
			\State $y_{idx}, idx\gets \min(A.y)$; 
			\While{$y_{idx}\leq y_{max}$}
			\State $p \gets A[idx]$; 
			\State Add $p$ to $B$;
			\State Remove points closer than $h(p)$ from $A$;
			\State Find nearest remaining points in candidates to the left and to the right of $p$;
			\State Add $5$ new points to the potential node set $A$, equispaced on the circular sector with center $p$ and a radius of $h(p)$, spanning from the nearest left to the nearest right point;
			\State $y_{idx}, idx\gets \min(A.y)$;
			\EndWhile
			\State\Return $B$
			\EndFunction
		\end{algorithmic}
	\end{algorithm}
	
	The original text did not provide the time complexity analysis, and the analysis was presented in a subsequent study  \cite{slakGenerationNodeDistributions2019a}. According to the analysis, the algorithm has a complexity of $O(NS)$, where $S$ is the maximum size of the potential node set, and $N$ denotes the number of generated nodes. 
	In practice, the running time of FF can be ignored compared to the expensive computational cost of training PINN.

	Since our subsequent experiments focus on time-dependent one-dimensional problems and a time-independent two-dimensional problem, where computational domains of the problems are essentially rectangles, FF is competent enough as the primary node generation method. More complex situations can also be handled by similar node generation frameworks. Due to the limited scope of the paper, we only consider 2D rectangular domains. We will discuss several potential improvements in the final section for higher-dimensional and more complex geometries.
	
	\section{Methods}
	We present a node generation method based on any given error map in the first subsection. In the second subsection, the method is then combined with the PINN, where PDE residuals are regarded as the error maps during the PINN training process.
	
	\subsection{Error-based Adaptive Node Generation with Memory}
	
	Suppose that a posteriori error map $e$ is given, where $e(x,y)$ represents an error indicator between an actual and computed value of some desired quantity at a point $(x,y)$. A larger absolute value of $e(x,y)$ indicates that the region near $(x,y)$ needs refinement. 
	On the other hand, we hope to incorporate a prior error map, through which we can build a memory mechanism that refines the regions where large error values have previously existed. Our motivation for the step will be illustrated in the numerical experiment part (Section \ref{sec:ac}). By introducing a prior error map $\tilde e$, we implement Algorithm \ref{alg:ff-ar} based on Algorithm \ref{alg:ff}.
	
	\begin{algorithm}
		\caption{Adaptive Refinement Node Positioning algorithm with FF}\label{alg:ff-ar}
		\begin{algorithmic}[1]
			\Require Box $[x_{min}, x_{max}]\times[y_{min},y_{max}]$,
			\Statex Error map $e: [x_{min}, x_{max}]\times[y_{min}, y_{max}]\rightarrow (-\infty,\infty)$,
			\Statex Prior error map $\tilde e: [x_{min}, x_{max}]\times[y_{min}, y_{max}]\rightarrow [0,1]$,
			\Statex Memory coefficient $\beta\in[0,1]$,
			\Statex Density ratio $r\in[1,\infty)$,
			\Statex Radius scaling parameter $s$.
			\Statex \textbf{Output} A set of points in the box and a prior error map for the next call.
			\Function{ARFF}{$x_{min}, x_{max}, y_{min}, y_{max}, e, \tilde e, \beta, r, s$}
			\State $\bar e\gets \max\left\{\frac{|e|-\inf_B |e|}{\sup_B|e|-\inf_B|e|+\varepsilon}, \beta \tilde e\right\}$ \Comment{$\varepsilon$ is a small positive number}
			\State $h\gets \left((1-\bar e)\times(1-1/\sqrt r)+1/\sqrt{r}\right)s$
			\State $pts\gets FF(x_{min}, x_{max}, y_{min}, y_{max}, h)$
			\State\Return $pts$, $\bar e$
			\EndFunction
		\end{algorithmic}
	\end{algorithm}
	
	
	Algorithm \ref{alg:ff-ar} requires two function parameters ($e, \tilde e$) and three real hyperparameters ($\beta, r, s$) as the inputs to control node generation. In subsequent uses, function variables are not directly kept. Instead, we use the nearest neighbor interpolation with values on a uniform grid to approximate the error map. Another possible option is to interpolate directly on the collocation point for approximation, replacing the uniform grid.
	
	
	\textbf{Standardize the error function}. Suppose $e$ is the error function as an input, the range of which can vary a lot in different applications. Therefore, values of the range are first standardized to $[0, 1)$. Combined with the standardized prior error function $\tilde e$, the nonlinear operator $\max$ is used to maintain sustained memory of large residual regions:
	\begin{align*}
		\bar e\gets \max\left\{\frac{|e|-\inf_B |e|}{\sup_B|e|-\inf_B|e|+\varepsilon}, \beta \tilde e\right\}
	\end{align*}
	The standardized error map $\bar e$ used in the current sampling depends on both the current error map $e$ and the prior error map $\tilde e$, where the memory coefficient $\beta$ controls the impact of $\tilde e$.
	In particular, the memory coefficient $\beta=0.0$ implies that the memory mechanism will not work. While $\beta=1.0$, the algorithm will never forget the areas where large residuals have existed before.
	Throughout all the experiments in this paper, the memory coefficient is not specially finetuned, and $\beta=0.9$ is applied to all cases.
	
	\textbf{Sampling radius function}. The sampling radius function $h$ directly controls the generation radius of the FF algorithm in each position, thereby controlling the local point density.
	The current standardized error map $\bar e$ partly determines the radius function $h$.
	The sampling radius at the large error position is small, thereby higher density; the sampling radius at the small error position is large, thereby lower density.
	The two sampling parameters ($r,s$) determine the eventual maximum and minimum generation radius in the rectangle. The ratio parameter $r$ controls the maximum to minimum radius ratio, thereby determining the ratio of the maximum to minimum density. The scaling parameter $s$ controls the overall scaling of radius. In the 2D plane, the maximum and  minimum sampling radius is estimated by
	\begin{align*}
		\max{h} = s,\quad \min{h}=s/\sqrt{r}.
	\end{align*}
	When the radius parameter $s$ is doubled, the density will reduce to a quarter of the original.

\begin{figure}[htbp]
		\centering
		\includegraphics[width=1\textwidth]{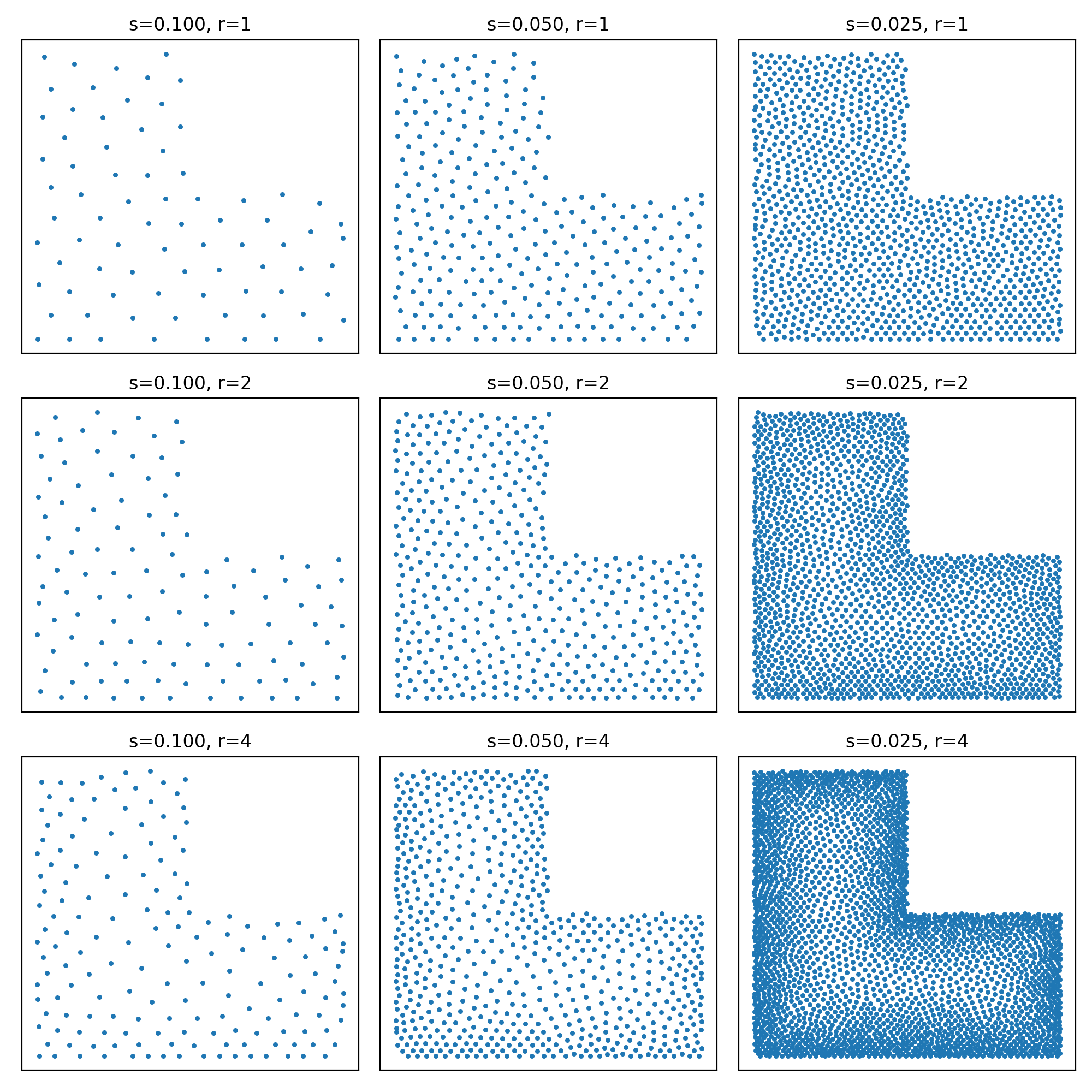}
		\caption{Nodal distribution in an L-shape domain. Sample nodes inside the L-shape according to $e(x,y):=1-SDF(x,y)$ under different scaling parameters $s$ and ratio parameters $r$. Each row has the same density distribution up to a constant factor, and the minimum distribution density in each column is the same.}
		\label{Lshape}
	\end{figure}

	For an intuitive warm-up, we use the signed distance function (SDF) of an L-shape to construct an error map to illustrate the effects of parameters $r$ and $s$.
	As is shown in Figure \ref{Lshape}, considering sampling on the L-shape inside a box domain $[0,1]\times[0,1]$, define $e:=1-SDF$, which implies high density near the boundary.
	Since Algorithm \ref{alg:ff} is designed for rectangles, we will sample nodes in the box domain and then remove nodes that lie outside of the L-shape.
	Figure \ref{Lshape} shows the sampling results under different parameters. We can see that $s$ determines the radius in the lowest density area. Keep $s$ unchanged, and then $r$ controls the maximum to minimum density ratio.
	The sampling radius function $h$ uses only two necessary hyperparameters ($r, s$), and is implemented in a simple weighted linear model (Line 3 in Algorithm \ref{alg:ff-ar}). Throughout all the experiments in this paper, we fix the ratio parameters $r$ to be $100$. The selection of $s$ will be discussed in the following subsection.
	
	\subsection{RANG for PINN}

	\textbf{Resample the collocation node set.}
	We define the resampling interval $I$, then the set will be re-generated every $I$ iterations. The collocation node set $\mathbf{N}_{pde}$ in Eq.\eqref{eq:pde_l} no longer stays unchanged. The PDE loss term Eq.\eqref{eq:pde_l} in the $i$-th iteration becomes
	\begin{align}
		L_{pde}(\theta)&=\frac{1}{\#\mathbf{M}_{\lfloor i/I\rfloor}}\sum_{(x,t)\in \mathbf{M}_{\lfloor i/I\rfloor}} |r_{pde}(t,x,\theta)|^2,
	\end{align}
	where $\mathbf{M}_{\lfloor i/I\rfloor}$ denotes the collocation point set used in the $i$-th iteration.
	The PDE residual $r_{pde}$ is employed as the error map in Algorithm \ref{alg:ff-ar}, which is applied to generate collocation nodes every $I$ iterations. Then Algorithm \ref{alg:ff-ar} is incorporated into the training process, and we obtain the Residual-based Adaptive Node Generation method (RANG) for PINN described in Algorithm \ref{alg:PINN-arff}.
	
	\textbf{Determine the scaling coefficient}. 
	Drastically varying the size of data sets may lead to a waste of storage resources such as RAM or video memory since we have to allocate enough space for the possible maximum size of the node set. Therefore we hope the number keeps roughly the same during training. However, when Algorithm \ref{alg:ff-ar} is applied to generate nodes, it is difficult to control the total number directly.
	
	To keep the size of collocation point sets nearly constant, we employ a loose bisection search for the scale parameter $s$ to generate a designated amount of points during training PINN, as shown in lines 10-16 in Algorithm \ref{alg:PINN-arff}.
	
	
	\textbf{Memory mechanism}. We note that refining a local region will decrease the PDE residual of the local region within several iterations. 
	However, once the local residual is decreased and the node density of the local region is reduced adaptively, the local PDE residual will increase again in a few iterations, and more nodes will be sampled in the region again. 
	The periodic increase and decrease of the local density will be repeated several times, and the oscillation might hamper the training efficiency. 
	After the memory mechanism is introduced, the training on the region will continue strengthening for a period of time even after the local residual is reduced. 
	An intuitive example is given in Section \ref{sec:ac}.
	
	\begin{algorithm}
		\caption{Residual-based Adaptive Node Generation for PINN (RANG)}\label{alg:PINN-arff}
		\begin{algorithmic}[1]
			\Require $s_{low}$, $s_{up}$ for bisection search of $s$,
			\Statex Maximal iterations $Maxiter$,
			\Statex Designated number of collocation points for PDE loss $N_{pde}$,
			\Statex Resampling interval $I$.
			\State Generate nodes for initial loss, boundary loss;
			\State Generate nodes for PDE loss with FF with a constant radius;
			\State $i\gets 0$
			\While{$i<Maxiter$}
			\If{$i\mod I=0$}
			\If{$i=0$}
			\State $e\gets 0$, $\tilde e\gets 0$
			\Else
			\State $e\gets\left|\mathcal{N}(t,x,\hat u(t,x;\theta))\right|$
			\EndIf
			\State $\bar s_{low}\gets s_{low}$, $\bar s_{up}\gets s_{up}$
			\State $s\gets (\bar s_{low}+\bar s_{up})/2$
			\While{$|\#\mathbf{M}_{\lfloor i/I\rfloor}-N_{pde}|/|N_{pde}|>0.05$ or $\bar s_{up}-\bar s_{low}<0.003$}
			\If{$\#\mathbf{M}_{\lfloor i/I\rfloor}<N_{pde}$}
			\State $\bar s_{up}\gets s$
			\Else
			\State $\bar s_{low}\gets s$
			\EndIf
			\State $s\gets (\bar s_{low}+\bar s_{up})/2$
			\State $\mathbf{M}_{\lfloor i/I\rfloor}, \bar e\gets ARFF(x_{min}, x_{max}, y_{min}, y_{max}, e, \tilde e, \beta, r, s)$
			\EndWhile
			\State $\tilde e\gets \bar e$
			\EndIf
			\State Compute the total loss $L$ in Eq.\eqref{eq:total_loss}.
			\State $\hat u\gets Adam(L)$\Comment{or other optimizers such as L-BFGS}
			\State $i\gets i+1$
			\EndWhile
		\end{algorithmic}
	\end{algorithm}
	
	\section{Results}
	In order to distinguish memoryles ($\beta=0.0$) and memorable ($\beta=0.9$) sampling methods in the numerical experiments, 
	let RANG represent Algorithm \ref{alg:PINN-arff} with $\beta=0.0$, and denote RANG-m for the case where $\beta=0.9$ holds, which is short for \textit{RANG with Memory mechanism}.
	
	This section employs nine sampling methods to generate collocation node sets. 
	We consider solving six different PDE equations, where the nonlinear equation includes a 1D Allen-Cahn equation, a 1D Schr\"odinger equation, and a 1D Korteweg-de Vries (KdV) equation, the linear equation includes a 1D wave equation, a 2D Poisson equation, and a 1D convection-diffusion equation. We start with the Allen-Cahn equation to illustrate the primary motivation of the memory mechanism. Then we observe the time-marching property of the RANG-m method in the wave equation. The remaining four PDEs will be tested in turn.
	
	Since the performance of PINNs may heavily depend on hyperparameters, the optimal choices for these hyperparameters are problem-dependent. 
	However, we focus on discussing the effects of the sampling methods on the results, and we use the following settings throughout the section without finetuning. 
	For all problems, fully connected networks are used and contain four hidden layers, each with $64$ neurons.
	The hyperbolic tangent $\tanh$ is used as the activation function for every layer except for the output layer.
	The Adam method is used to train PINNs, with the learning rate fixed to $0.001$.
	

	
	Due to the FF method being proposed for generating nodes in a two-dimensional rectangle, RANG based on the FF method is only suitable for two-dimensional problems. 
	Note that the domains of time-dependent one-dimensional PDEs are essentially two-dimensional. 
	Therefore, the following six experiments consider the collocation point sampling methods in rectangles, where nodes are generated in a square of $[0,1]\times[0,1]$ and mapped to the rectangle with an affine transformation.
	For all examples, the following nine different sampling methods are taken into consideration:
	
	\begin{enumerate}
		\item \textbf{Random}. Nodes are independent and identically distributed samples on the domain, obeying the uniform distribution.
		\item \textbf{Random-R}. Resample nodes using the Random method every $I$ iterations.
		\item \textbf{Hammersley}. The Hammersely method with $p=2$ is used for node generation.
		\item \textbf{LHS}. Latin hypercubic sampling is employed.
		\item \textbf{LHS-R}. Resample nodes using LHS every $I$ iterations.
		\item \textbf{FF}. Sample nodes with FF where the radius function is constant on the rectangular domain, and the radius constant is determined by a bi-section search.
		\item \textbf{FF-R}. Resample nodes using FF every $I$ iterations.
		\item \textbf{RANG}. Residual-based Adaptive Node Generation with the memory coefficient $\beta=0.0$ is employed. The method resamples nodes every $I$ iterations based on PDE residuals.
		\item \textbf{RANG-m}. Residual Adaptive Node Generation with the memory coefficient $\beta=0.9$ is employed. The method also resamples nodes every $I$ iterations.
	\end{enumerate}

	The above nine sampling methods can be divided into two categories: the first category is the one-time sampling method, including Random, Hammersley, FF, and LHS, of which the sampling nodes are generated before training and will be used throughout the entire training process; 
	the second is the resampling method including the remaining five approaches. 
	
	In order to reduce the effects of other stochastic factors on the results, we only consider sampling methods applied to PDE-informed loss terms. For boundaries and initial conditions, by default, equispaced nodes are used. 
	Of course, for more complex boundaries, adaptive sampling should also be considered, and we will extend this method in future research.
	
	For all examples, the accuracy of $\hat u$ is evaluated by the mean squared error (MSE) after training for a fixed number of iterations:
	\begin{align*}
		MSE(\hat u)=\frac{1}{\#\mathbf{G}}\sum_{i\in \mathbf{G}} \|u(t_i, x_i)-\hat u(t_i, x_i)\|^2
	\end{align*}
	where $u$ is the reference solution, and $\mathbf{G}$ is the testing set generated by a uniform grid.
	
	All the source code is implemented with python and PyTorch, available online (\url{https://github.com/weipengOO98/rang_pinn}).
	
	\subsection{Allen-Cahn Equation}\label{sec:ac}
	The first nonlinear equation we solve is the Allen-Cahn equation. Consider the following form:
	\begin{align*}
		\frac{\partial u}{\partial t} - 0.0001\frac{\partial^2 u}{\partial x^2} + 5u^3 - 5u=0, \\
		\quad(t,x)\in[0,1]\times[-1,1],\\
		u(t,-1)=u(t,1)=-1,\\
		u(0,x)=x^2\cos(\pi x).
	\end{align*}
	The neuron numbers for each layer are $[2,64,64,64,64,1]$.
	The maximum iteration number is set to $50,000$, and the resampling interval $I$ is set to $1,000$. The number of replicates is $30$. 
	The initial condition loss term is
	\begin{align*}
		L_0 = \frac{1}{\#\mathbf{N}_0}\sum_{x_0\in\mathbf{N}_0}\left(|u(0,x_0)-x_0^2\cos(\pi x_0)|^2+|u_x(0,x_0)|^2\right).
	\end{align*}
	The boudnary condition loss term is
	\begin{align*}
		L_b = \frac{1}{\# \mathbf{N}_b}\sum_{t_b\in\mathbf{N}_b}\left(|u(t_b,-1)+1|^2+|u(t_b,1)+1|^2\right).
	\end{align*}
	The designated number of collocation points $N_{pde}$ for the PDE loss term is set to $1,000$. The PDE loss term becomes 
	\begin{align*}
		L_{pde} = \frac{1}{\#\mathbf{M}_{j}}\sum_{(\tilde x,\tilde t)\in \mathbf{M}_{j}}\left(\frac{\partial u(\tilde t,\tilde x)}{\partial t} - 0.0001\frac{\partial^2 u(\tilde t, \tilde x)}{\partial x^2} + 5u^3(\tilde t, \tilde x) - 5u(\tilde t, \tilde x)\right)^2,
	\end{align*}
	where $\mathbf{M}_j$ denotes the $j$-th node set employed, which is used from the $j\times10^3$-th iteration to the $(j+1)\times 10^3$-th iteration. For the resampling methods, the collocation point set $\mathbf{M}_j$ varies every $I$ iterations ($j=0,1,2,\cdots$). For the non-resampling methods, the PDE loss collocation point set $\mathbf{M}_j=\mathbf{M}_0$ holds, which stays unchanged. $\#\mathbf{M}_j=N_{pde}$ holds for Random, Random-R, LHS, LHS-R and Hammersley, and $\#\mathbf{M}_j\approx N_{pde}$ holds for FF, FF-R, RANG and RANG-m. 
	The total loss is a weighted sum of the initial value loss term, the boundary value loss term, and the PDE loss term.
	We choose $w_0=2.0$, $w_b=2.0$, $w_{pde}=0.2$ in Eq.\eqref{eq:total_loss} for better enforcing the initial and boundary conditions:
	\begin{align*}
		L=2L_0+2L_b+0.2L_{pde}.
	\end{align*}
	
\begin{figure}[htbp]
		\centering
		\includegraphics[width=1\textwidth]{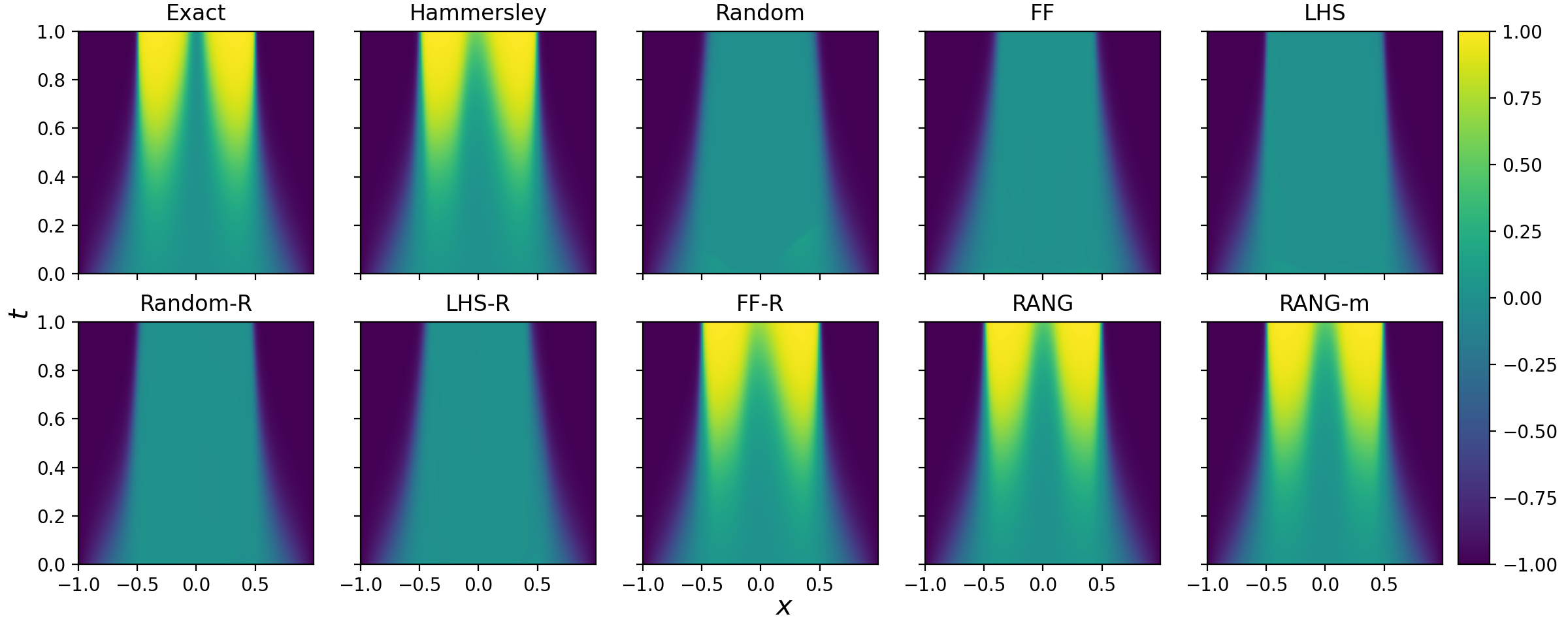}
		\caption{Solutions for the Allen-Cahn equation. The top left image exhibits the reference solution of the Allen-Cahn equation. The remaining nine images show the prediction results of nine sampling methods in one experiment. The results might be different in another run due to the randomness of sampling and training.}
		\label{fig:essemble_ac}
	\end{figure}

	Variants of PINNs have solved the Allen-Cahn equation in several studies.
	However, as reported by Zhao \cite{zhaoSolvingAllenCahnCahnHilliard2021}, the vanilla PINN fails to solve the equation.
	Figure \ref{fig:essemble_ac} depicts the reference solution and the results of nine different sampling methods.
	
\begin{figure}[htbp]
		\centering
		\includegraphics[width=1\textwidth]{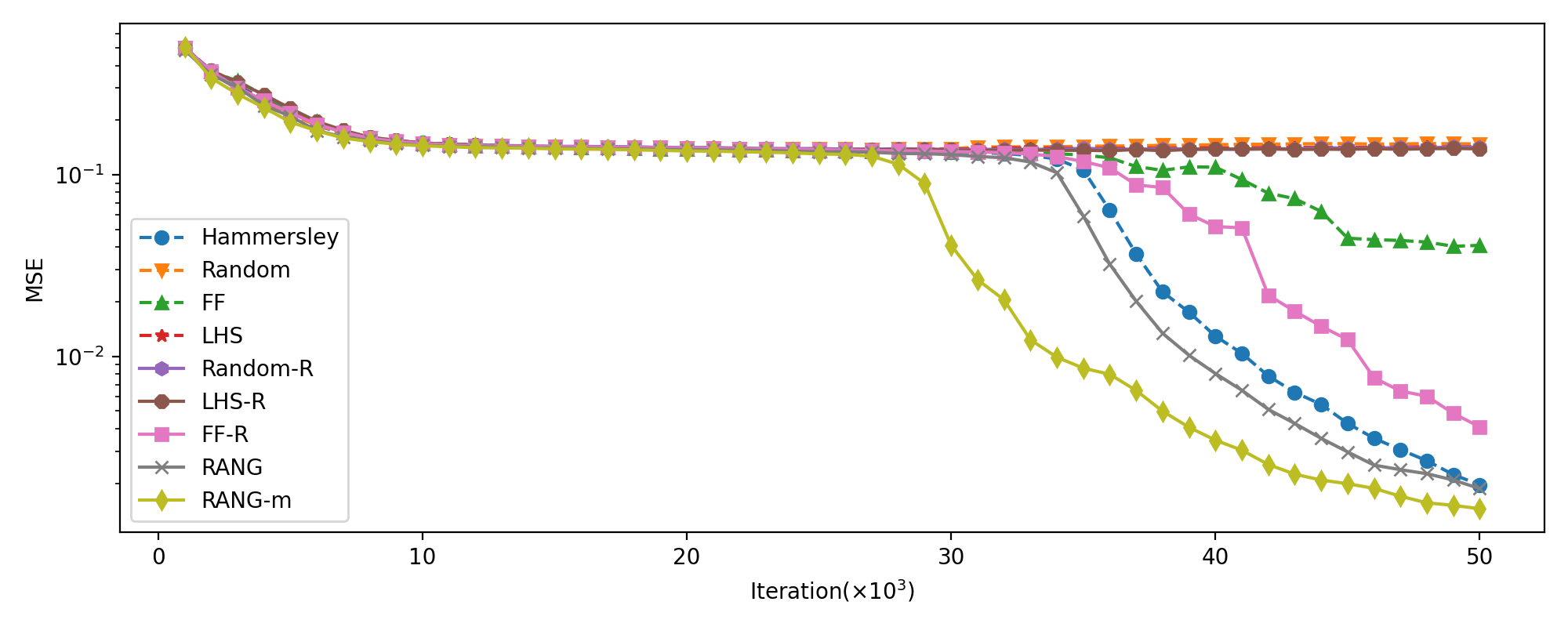}
		\caption{MSE values in iteration for the Allen-Cahn equation. This figure shows the curve of the median of MSE corresponding to the nine different methods. Note that four methods (Random/LHS/Random-R/LHS-R) are almost overlapped, and there is nearly no decrease in MSE. MSE values of the remaining methods decline, which might benefit from resampling and the local quasi-uniform properties.}
		\label{fig:ac_it}
	\end{figure}

\begin{figure}[htbp]
		\centering
		\includegraphics[width=1\textwidth]{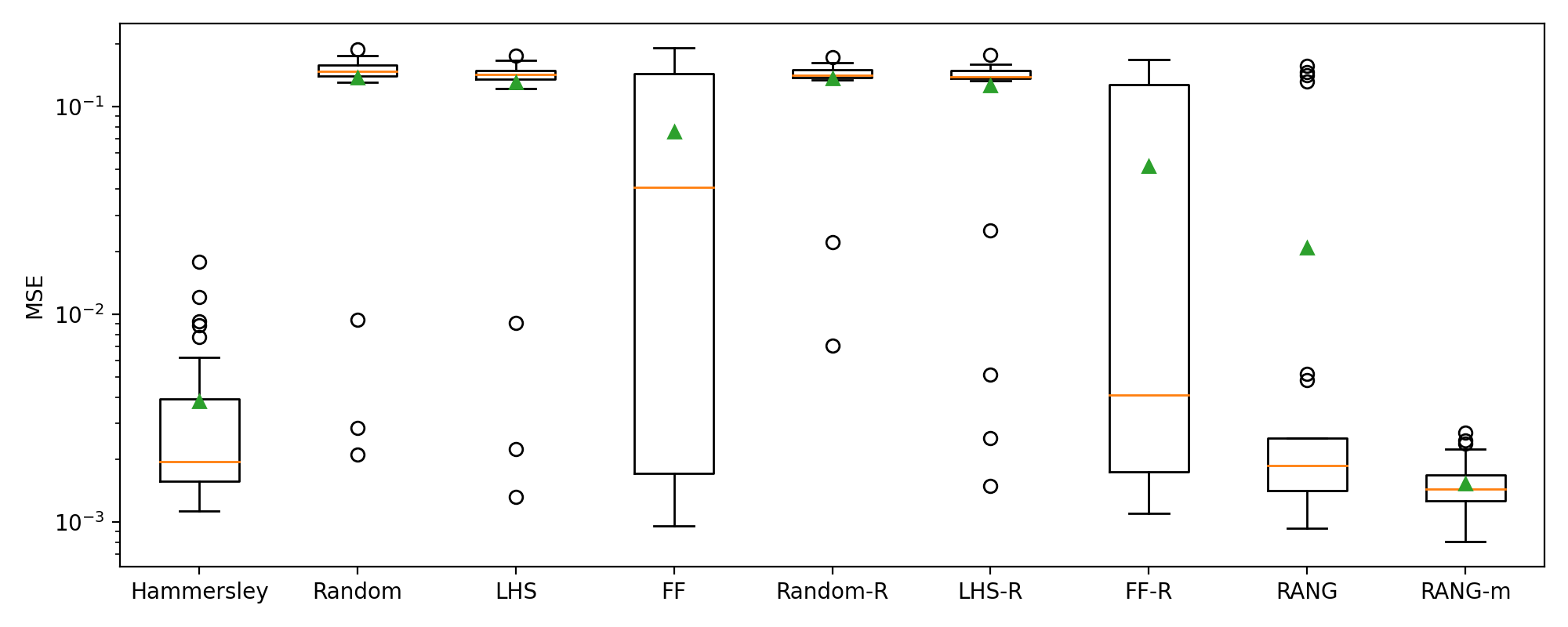}
		\caption{Box plot of MSEs for Allan-Cahn equation considering different node generation strategies. The plot represents median and 25, 75\% interquartile and extreme values. The mean values are marked with triangles.}
		\label{fig:ac_error}
	\end{figure}

	As shown in Figure \ref{fig:ac_it}, it tends to saturate for an extended period in the training process of PINNs. After a lengthy computation, the MSE may fall drastically, where RANG-m has the fastest decline.
	After $50,000$ iteration, statistics of MSEs obtained by nine sampling methods are shown in Figure \ref{fig:ac_error} and Table \ref{tab:ac}. All statistics indicate that RANG-m achieves the best performance. Consistent with the previous result \cite{dasStateoftheArtReviewDesign2022}, the Hammersley sampling performs best among the four non-resampling methods. In this example, we can also see that RANG-m is more stable than RANG.
	
\begin{figure}[htbp]
		\centering
		\includegraphics[width=1\textwidth]{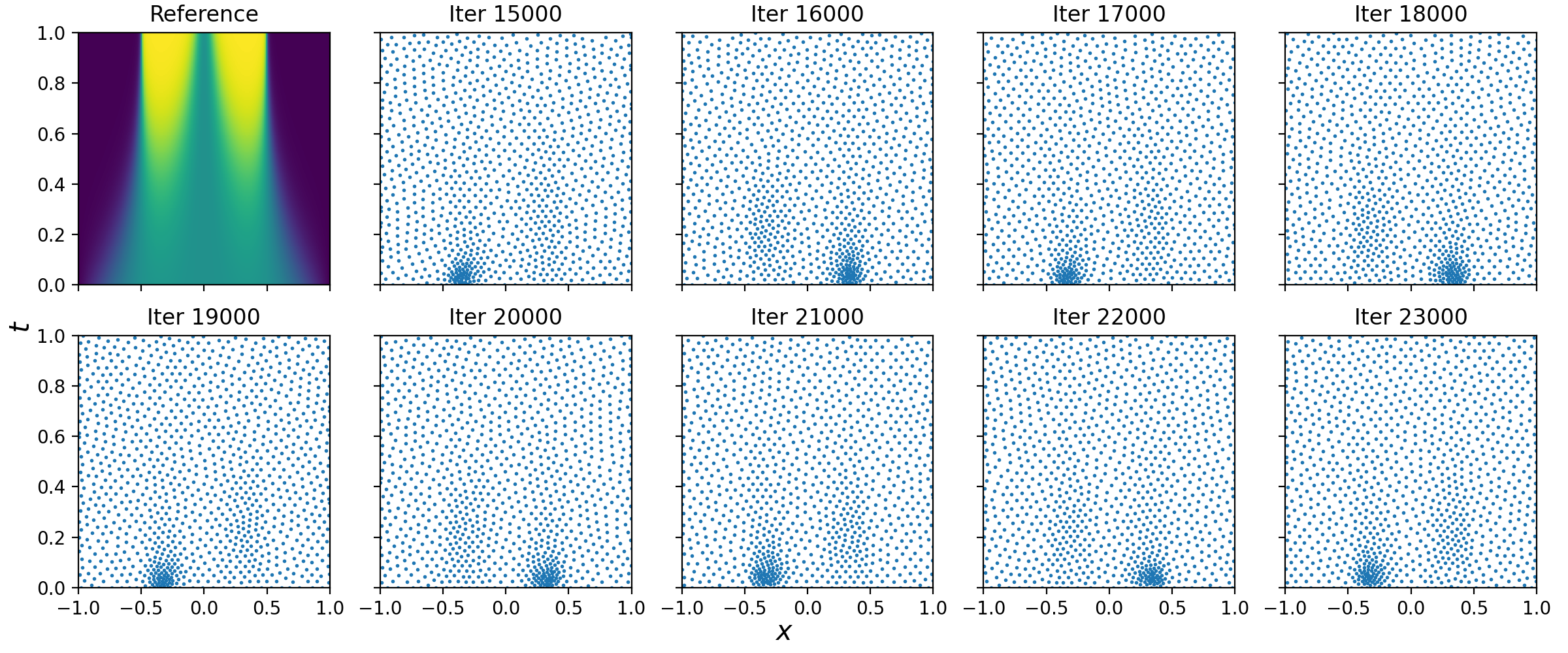}
		\caption{Adaptive nodal distribution via RANG. The top left image is a reference solution for the Allen-Cahn equation. The remaining nine images depict the RANG collocation nodes from 15,000-th to 23,000-th iteration. An interesting phenomenon is the periodic refinement of two different regions. This oscillation may hamper convergence speed.}
		\label{fig:jump0}
	\end{figure}

\begin{figure}[htbp]
		\centering
		\includegraphics[width=1\textwidth]{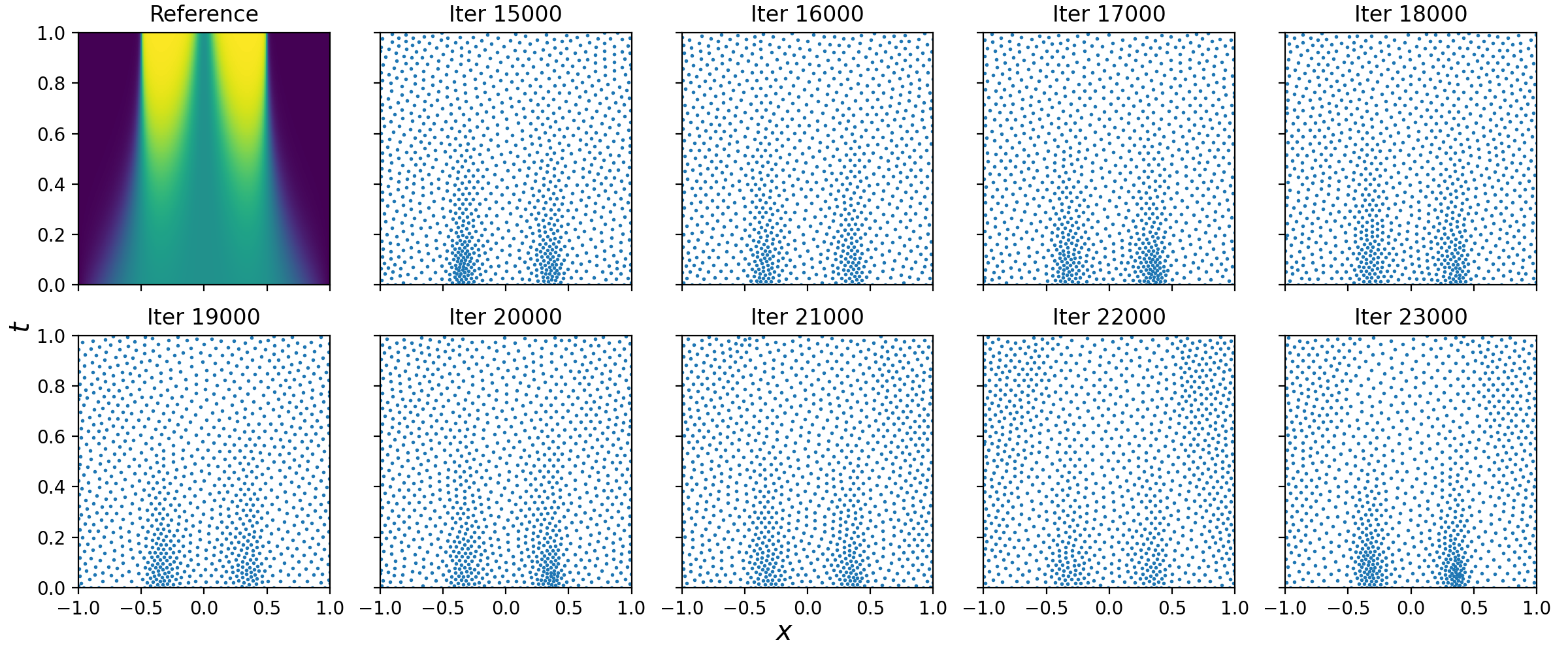}
		\caption{Adaptive nodal distribution via RANG-m. The top left image is a reference solution for the Allen-Cahn equation. The remaining nine images depict the RANG-m collocation nodes from 15,000-th to 23,000-th iteration. The memory mechanism can continuously refine the region that previously had a large residual for a period of time.}
		\label{fig:jump9}
	\end{figure}

	\begin{table}
		\begin{tabular}{l|rrrrrr}
			\hline
			{} &      mean &       min &       25\% &       50\% &       75\% &       max \\
			\hline
			Hammersley & 3.814E-03 & 1.135E-03 & 1.571E-03 & 1.959E-03 & 3.922E-03 & 1.788E-02 \\
			Random     & 1.382E-01 & 2.110E-03 & 1.403E-01 & 1.480E-01 & 1.589E-01 & 1.891E-01 \\
			LHS        & 1.310E-01 & 1.317E-03 & 1.356E-01 & 1.425E-01 & 1.495E-01 & 1.755E-01 \\
			FF         & 7.559E-02 & 9.563E-04 & 1.718E-03 & 4.085E-02 & 1.443E-01 & 1.910E-01 \\
			Random-R   & 1.366E-01 & 7.051E-03 & 1.376E-01 & 1.410E-01 & 1.501E-01 & 1.730E-01 \\
			LHS-R      & 1.265E-01 & 1.491E-03 & 1.362E-01 & 1.390E-01 & 1.486E-01 & 1.777E-01 \\
			FF-R       & 5.175E-02 & 1.097E-03 & 1.741E-03 & 4.086E-03 & 1.274E-01 & 1.688E-01 \\
			RANG       & 2.094E-02 & 9.378E-04 & 1.416E-03 & 1.869E-03 & 2.526E-03 & 1.574E-01 \\
			RANG-m     & \textbf{1.537E-03} & \textbf{8.052E-04} & \textbf{1.266E-03} & \textbf{1.448E-03} & \textbf{1.690E-03} & \textbf{2.702E-03} \\
			\hline
		\end{tabular}
		\caption{The statistics of MSE for the Allen-Cahn equation considering different node generation strategies over 30 replicates. The optimal value is marked in bold for each statistic. The RANG-m method performs best in various statistics.}
		\label{tab:ac}
	\end{table}
	Next we will show the motivation for introducing the memory mechanism.
	Figure \ref{fig:jump0} shows that the RANG method refines two local regions during training (from 15,000-th to 23,000-th iteration).
	The collocation point set is resampled every $1,000$ iterations. It can be seen that RANG repeatedly refines two regions in turn. When region A is refined, the local PDE residual on region B will again increase, such that refinement is again required on region B. We propose the memory mechanism, which continues refining both A and B such that the periodic oscillation is prevented. As is shown in Figure \ref{fig:jump9}, the refinement seems steady during the training, which might benefit the stability of the final results of RANG-m.
	
	\subsection{1D Wave Equation}
	The first linear PDE equation we discuss is a one-dimensional wave equation that describes the deflection of a linear string over time and space.
	Given an interval $[-l,l]$ with fixed endpoints ($l=4$), the governing PDE equation with the initial and boundary conditions for the wave equation is given by
	\begin{align*}
		\frac{\partial^2 u}{\partial t^2}&-3\frac{\partial^2 u}{\partial x^2}=0,\\
		u(0,x)&=\frac{1}{\cosh(2x )}-\frac{0.5}{\cosh(2(x-2l))}-\frac{0.5}{\cosh(2(x+2l))},\\
		\frac{\partial u}{\partial t}(0,x)&=0,\\
		u(t,-l)&=u(t,l)=0,
	\end{align*}
	where the exact solution within the range $t\in[0,6]$ is
	\begin{align*}
		u^{ref}(x,t)=&\frac{0.5}{\cosh(2(x+\sqrt{3}t))}-\frac{0.5}{\cosh(2(x-2l+\sqrt{3}t))}\\
		&+\frac{0.5}{\cosh(2(x-\sqrt{3}t))}-\frac{0.5}{\cosh(2(x+2l-\sqrt{3}t))},
	\end{align*}
	which describes a pulse propagating on both sides and reflects at the boundary. 
	The neuron number for each layer is set to $[2,64,64,64,64,1]$, respectively.
	The initial condition loss consists of the initial position and velocity loss:
	\begin{align*}
		L_0 = \frac{1}{\#\mathbf{N}_0}\sum_{x_0\in\mathbf{N}_0}\left(|u(0,x_0)-u^{ref}(0,x_0)|^2+|u_x(0,x_0)|^2\right).
	\end{align*}
	The Dirichlet boundary loss term is
	\begin{align*}
		L_b = \frac{1}{\#\mathbf{N}_b}\sum_{t_b\in\mathbf{N}_b}\left(|u(t_b,-l)|^2+|u(t_b,l)|^2\right)
	\end{align*}
	The PDE loss term is given by
	\begin{align*}
		L_{pde} = \frac{1}{\# \mathbf{M}_{j}}\sum_{(\tilde x,\tilde t)\in \mathbf{M}_{j}}\left( \frac{\partial^2 u(\tilde t,\tilde x)}{\partial t^2}-3\frac{\partial^2 u(\tilde t,\tilde x)}{\partial x^2}\right)^2
	\end{align*}
	where $\mathbf{M}_j$ denotes the $j$-th node set employed. Since we set the resampling interval $I=1,000$, the collocation point set $M_j$ changes every $1,000$ iterations for the five resampling methods ($j=0,1,2,\cdots$). For the four non-resampling methods, the PDE loss collocation point set $\mathbf{M}_j=\mathbf{M}_0$ holds, which stays unchanged during the training.
	
	The total loss is a weighted sum of the initial value loss term, the boundary value loss term, and the PDE loss term. We also choose $w_0=2.0, w_b=2.0, w_{pde}=0.2$ for better enforcing the initial and boundary conditions:
	\begin{align*}
		L=2L_0+2L_b+0.2L_{pde}.
	\end{align*}
	
\begin{figure}[htbp]
		\centering
		\includegraphics[width=1\textwidth]{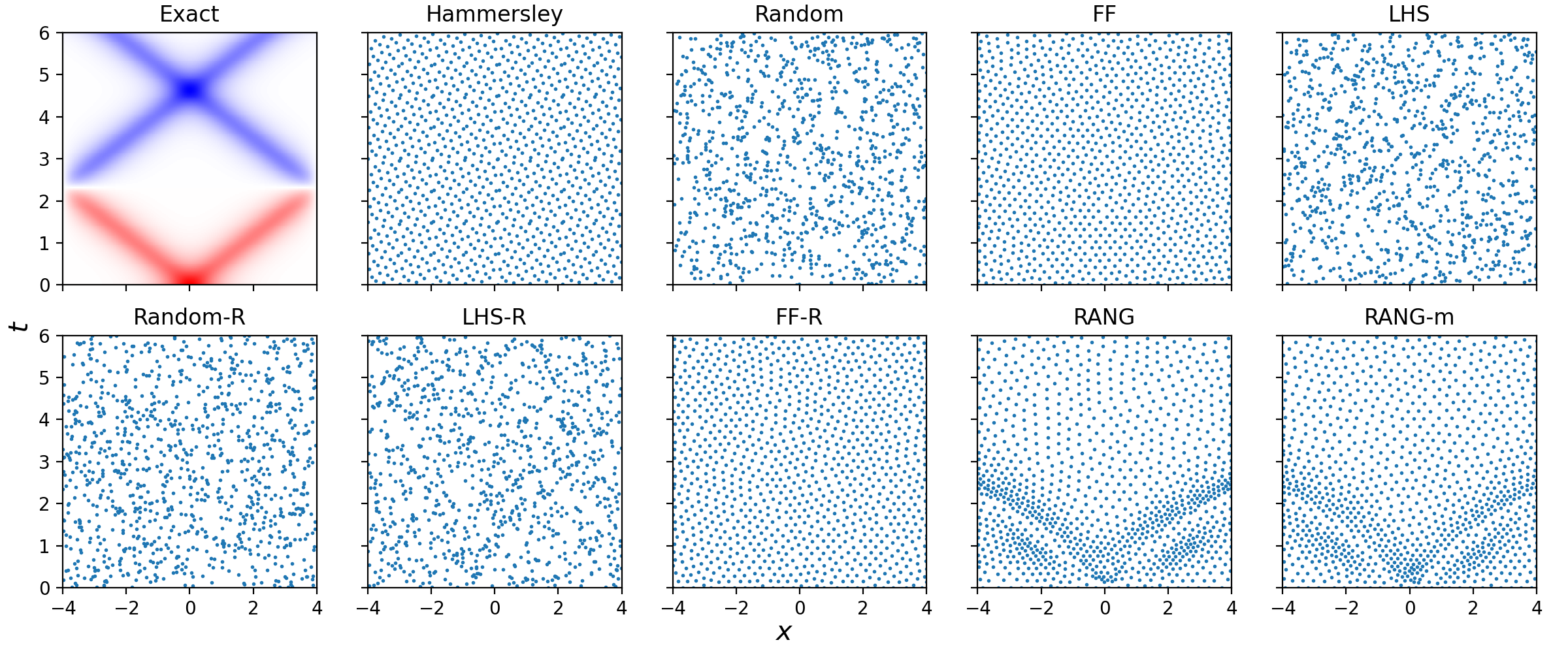}
		\caption{Nodal distribution for the wave equation at the 3,000-th iteration. The top left image shows the exact solution. The remaining nine images depict $\mathbf{M}_3$ at iteration 3,000 for each approach. The four node sets on the first row remain unchanged in the entire iteration process. The second row displays the five resampling methods that resample nodes every 1,000 iterations. RANG and RANG-m in the bottom right images refine the region of the fluctuation propagation trajectory in early iterations.}
		\label{fig:scatter_exp}
	\end{figure}

	Set the maximum iterations to be $15,000$, and the number of replicates is $50$. The designated capacity of collocation point set is $N_{pde}=1000$. 
	Figure \ref{fig:scatter_exp} displays the collocation point set $\mathbf{M}_3$ which is used from 3,000-th to 4,000-th iteration. Hammersley, FF, and FF-R exhibit low-discrepancy and their nodes are more evenly spread on the domain than Random. The nodes generated by RANG and RANG-m look locally regular, and their local density varies adaptivly according to the current PDE residual.
	
\begin{figure}[htbp]
		\centering
		\includegraphics[width=1\textwidth]{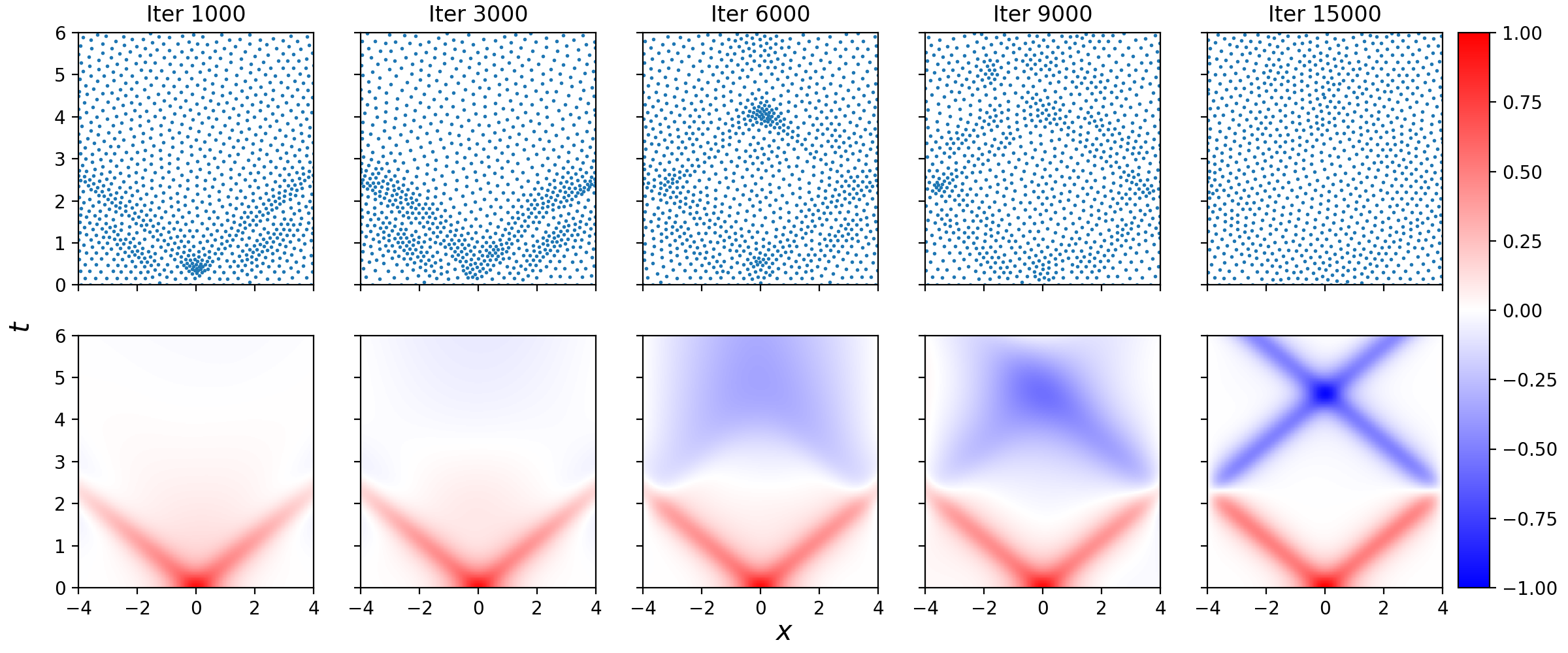}
		\caption{
			Nodal distributions and corresponding predicted solutions at different iterations. Each column displays the collocation set node and prediction of the RANG-m method at the same iteration. As the iteration goes on, the sampling nodes are gradually distributed evenly along the time axis.}
		\label{fig:time_marching}
	\end{figure}

\begin{figure}[htbp]
		\centering
		\includegraphics[width=1\textwidth]{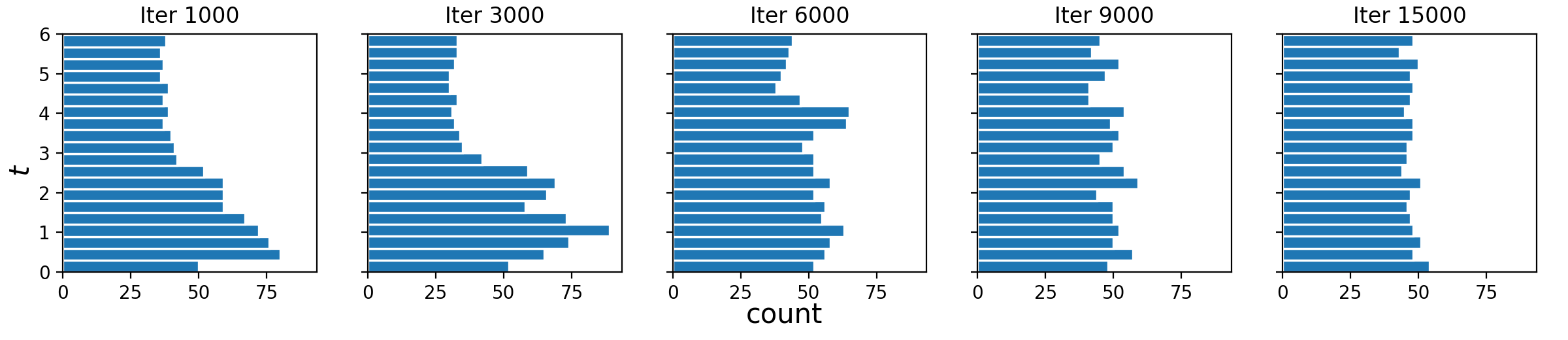}
		\caption{
			Histogram of nodes along the time axis. In the initial iteration stage, the sampling point is concentrated near $t=0$. The collocation points are gradually distributed evenly on the whole region with iteration going on.}
		\label{fig:time_marching_hist}
	\end{figure}

	Figure \ref{fig:time_marching} depicts the collocation nodes generated by RANG-m in different iterations.
	Figure \ref{fig:time_marching_hist} shows the frequency histogram of the sample point along the time axis. 
	It is shown that the sampling nodes are concentrated near $t=0$ in the early iterations.
	As the iteration goes on, the number of samples in subsequent regions gradually increases.
	After the training ends, the distribution of sampling points in the time axis also becomes uniform. The RANG method automatically increases the weight of the initial area, which may contribute to the training of PINNs.
	In this problem, the RANG method has a certain time-marching feature, which conforms to the causal rules of evolutionary equations. 
	
\begin{figure}[htbp]
		\centering
		\includegraphics[width=1\textwidth]{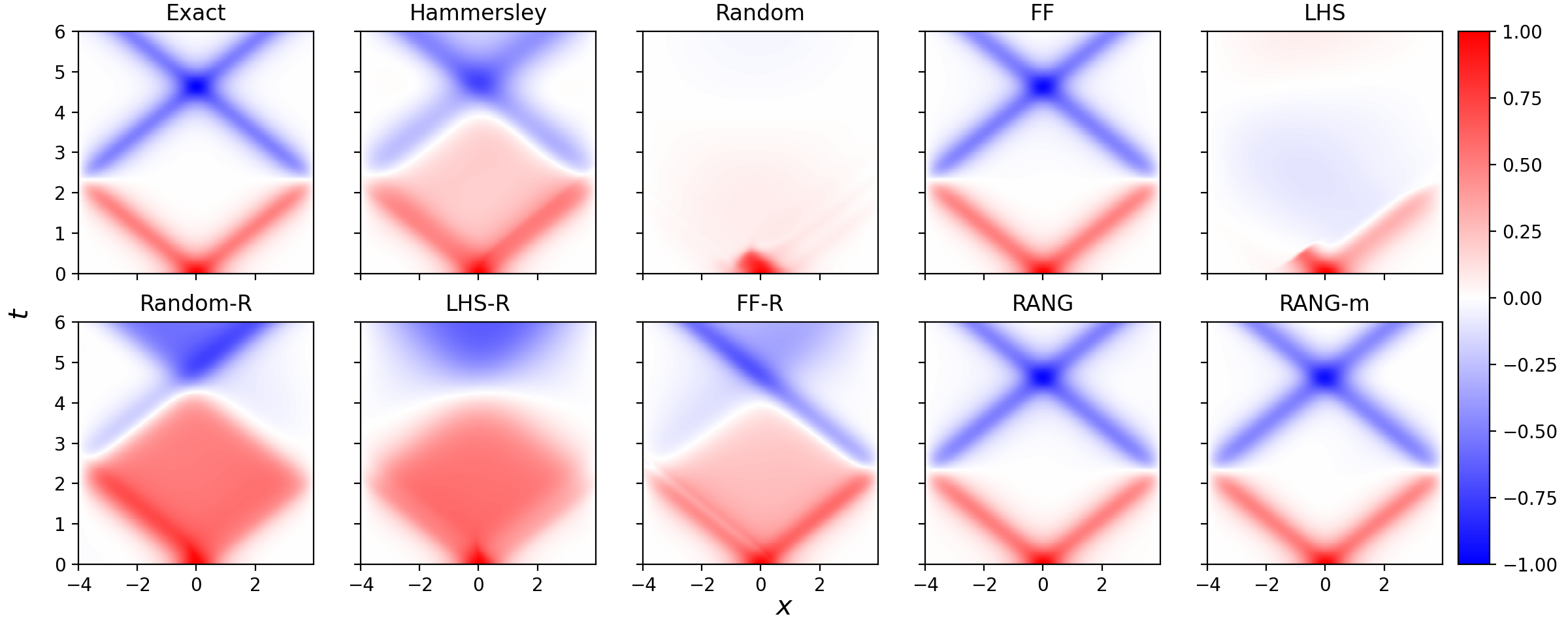}
		\caption{Solutions for the wave equation. The top left image exhibits the reference solution of the wave equation. The remaining nine images show the prediction results of nine sampling methods in one experiment. The results might be different in another run due to the randomness of sampling and training.}
		\label{fig:essemble_wave}
	\end{figure}

	Figure \ref{fig:essemble_wave} depicts the reference solution and the results of nine different sampling methods after $15,000$ iterations.
	
\begin{figure}[htbp]
		\centering
		\includegraphics[width=1\textwidth]{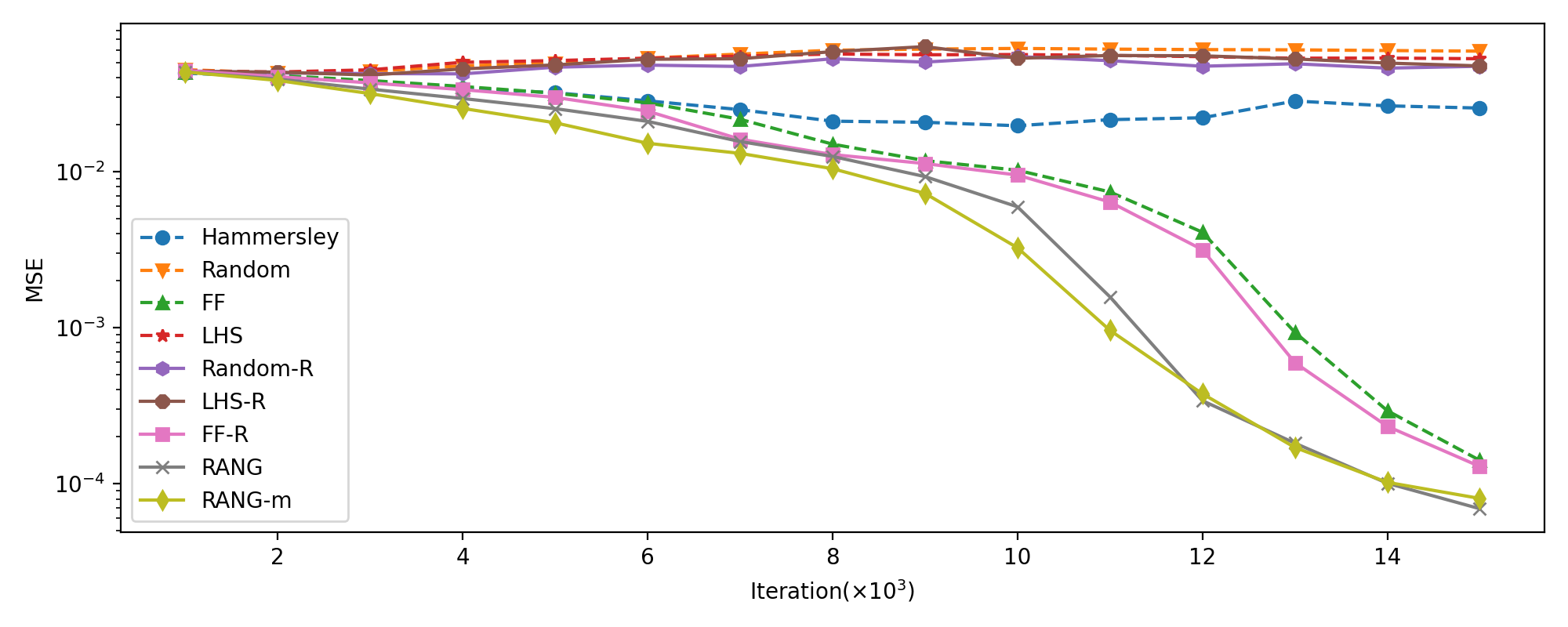}
		\caption{MSE values in iteration for the wave equation. This figure shows the curve of the median of MSE corresponding to the nine different methods.}
		\label{fig:wave_it}
	\end{figure}

	Figure \ref{fig:wave_it} shows that FF sampling has the best performance among the non-resampling methods, even better than Hammersley. In this example, the four FF based methods (FF, FF-R, RANG, RANG-m) all perform well.
	
\begin{figure}[htbp]
		\centering
		\includegraphics[width=1\textwidth]{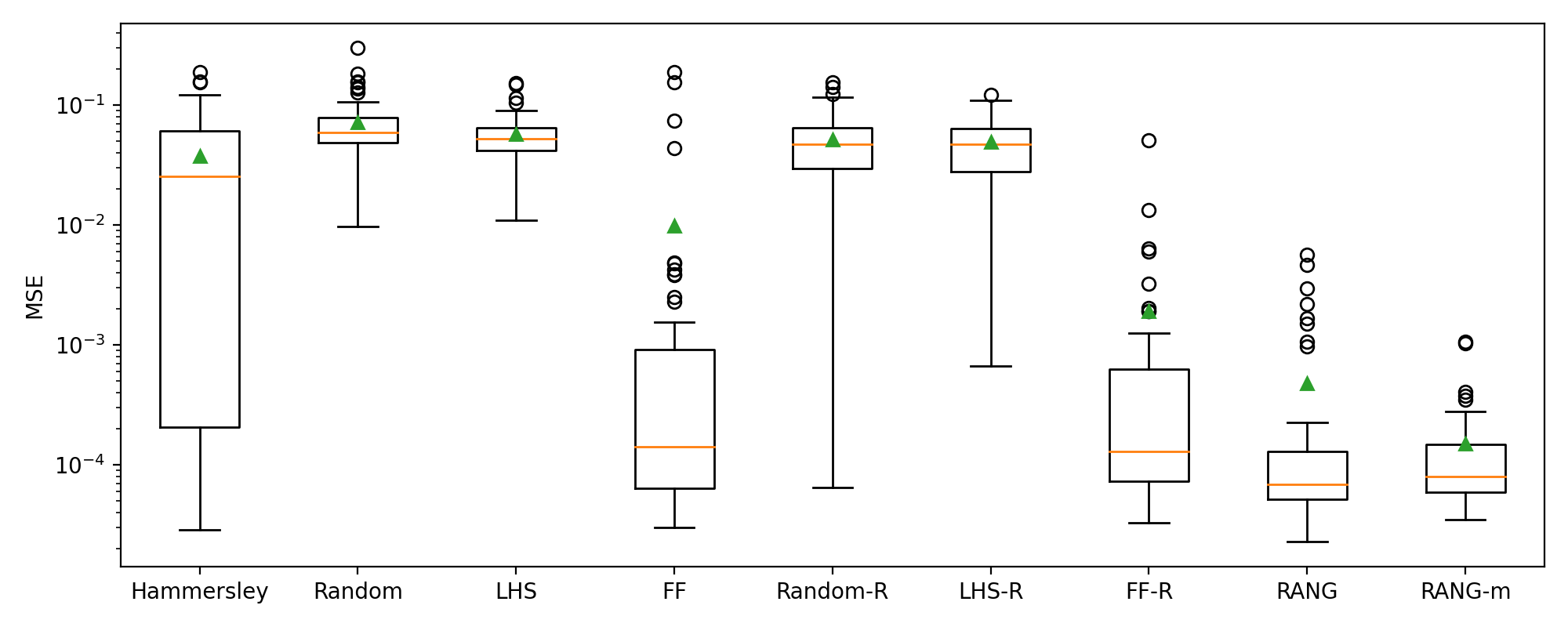}
		\caption{Box plot of MSEs for the wave equation considering different node generation strategies. The plot represents median and 25, 75\% interquartile and extreme values. The mean values are marked with triangles.}
		\label{fig:wave_error}
	\end{figure}

	Figure \ref{fig:wave_error} and Table \ref{tab:wave} display several statistics about MSE after 15,000 iterations with 50 replicates. Although the medians of RANF, FF, FF-R, and RANG-m are relatively close, the average value of RANG-m is much smaller, which implies the method may avoid poor convergence. The stability of RANG-m might benefit from the memory mechanism.
	
	\begin{table}
		\begin{tabular}{l|rrrrrr}
			\hline
			{} &      mean &       min &       25\% &       50\% &       75\% &       max \\
			\hline
			Hammersley & 3.793E-02 & 2.872E-05 & 2.066E-04 & 2.547E-02 & 6.058E-02 & 1.884E-01 \\
			Random     & 7.248E-02 & 9.773E-03 & 4.889E-02 & 5.897E-02 & 7.826E-02 & 2.988E-01 \\
			LHS        & 5.765E-02 & 1.101E-02 & 4.159E-02 & 5.274E-02 & 6.480E-02 & 1.535E-01 \\
			FF         & 9.878E-03 & 3.009E-05 & 6.422E-05 & 1.412E-04 & 9.143E-04 & 1.876E-01 \\
			Random-R   & 5.155E-02 & 6.457E-05 & 2.958E-02 & 4.719E-02 & 6.474E-02 & 1.560E-01 \\
			LHS-R      & 4.947E-02 & 6.648E-04 & 2.801E-02 & 4.744E-02 & 6.340E-02 & 1.220E-01 \\
			FF-R       & 1.910E-03 & 3.302E-05 & 7.295E-05 & 1.290E-04 & 6.270E-04 & 5.092E-02 \\
			RANG       & 4.790E-04 & \textbf{2.289E-05} & \textbf{5.153E-05} & \textbf{6.919E-05} & \textbf{1.289E-04} & 5.683E-03 \\
			RANG-m     & \textbf{1.508E-04} & 3.479E-05 & 5.878E-05 & 8.051E-05 & 1.485E-04 & \textbf{1.069E-03} \\
			\hline
		\end{tabular}
		\caption{
			The statistics of MSE for the wave equation considering different node generation strategies over 50 replicates. The optimal value is marked in bold for each statistic. RANG-m performs best in the \textit{mean} column.}
		\label{tab:wave}
	\end{table}

	\subsection{1D Sch\"odinger Euqation}
	The 1D Sch\"odinger equation is the second nonlinear equation in our experiments. Consider the following complex-valued PDE:
	\begin{align*}
		\mathrm{i} \frac{\partial u}{\partial t}+\frac{1}{2}\frac{\partial^2 u}{\partial x^2}+|u|^2\cdot u=0,\\
		u(0,x)=2\mathrm{sech}(x),\\
		u(t,-5)=u(t,5),\\
		\frac{\partial u}{\partial x}(t,-5)=\frac{\partial u}{\partial x}(t,5),
	\end{align*}
	where
	\begin{align*}
		(x,t)\in \Omega = [-5,5]\times[0,\pi/2].
	\end{align*}
	The solution $u$ is complex-valued, which can be equivalently represented by two real-valued functions, so the number of output layer neurons of the neural network $\hat u$ is $2$. 
	The number of neurons in each layer is $[2, 64, 64, 64, 64, 2]$, respectively.
	
	The initial condition loss term is
	\begin{align*}
		L_0 = \frac{1}{\#\mathbf{N}_0}\sum_{x_0\in\mathbf{N}_0}|u(0,x_0)-2\cdot\mathrm{sech}(x_0)|^2.
	\end{align*}
	The boundary condition loss term is
	\begin{align*}
		L_b = \frac{1}{\#\mathbf{N}_b}\sum_{t_b\in\mathbf{N}_0}\left(|u(t_b,-5)-u(t_b, 5)|^2+|u_x(t_b,-5)-u_x(t_b, 5)|^2\right),
	\end{align*}
	where $u(t,x)=v(t,x)+\mathrm{i}\cdot w(t,x)$, $v,w$ are two real-valued functions. The complex derivative $u_x$ is defined as
	\begin{align*}
		u_x:=\frac{\partial v}{\partial x}+\mathrm{i}\frac{\partial w}{\partial x}.
	\end{align*}
	The governing PDE loss is
	\begin{align*}
		L_{pde} = \frac{1}{\# \mathbf{M}_j}\sum_{(\tilde x,\tilde t)\in \mathbf{M}_{j}}\left(\left|\mathrm{i} \frac{\partial u}{\partial t}+\frac{1}{2}\frac{\partial^2 u}{\partial x^2}+|u|^2\cdot u\right|^2\right).
	\end{align*}
	
	For the non-resampling methods, the PDE loss collocation point set $\mathbf{M}_j=\mathbf{M}_0$ holds,  which stays unchanged. The total loss is a weighted sum of the initial value loss term, the boundary value loss term and the PDE loss term.
	We also choose $w_0=2.0, w_b=2.0, w_{pde}=0.2$ for better enforcing the initial and boundary conditions:
	\begin{align*}
		L=2L_0+2L_b+0.2L_{pde}.
	\end{align*}
	The numerical reference solution could be obtained via Fourier spectral semi-discrete scheme with Runge-Kutta integration. We refer to the numerical solution provided by \cite{raissiPhysicsinformedNeuralNetworks2019}.
	
\begin{figure}[htbp]
		\centering
		\includegraphics[width=1\textwidth]{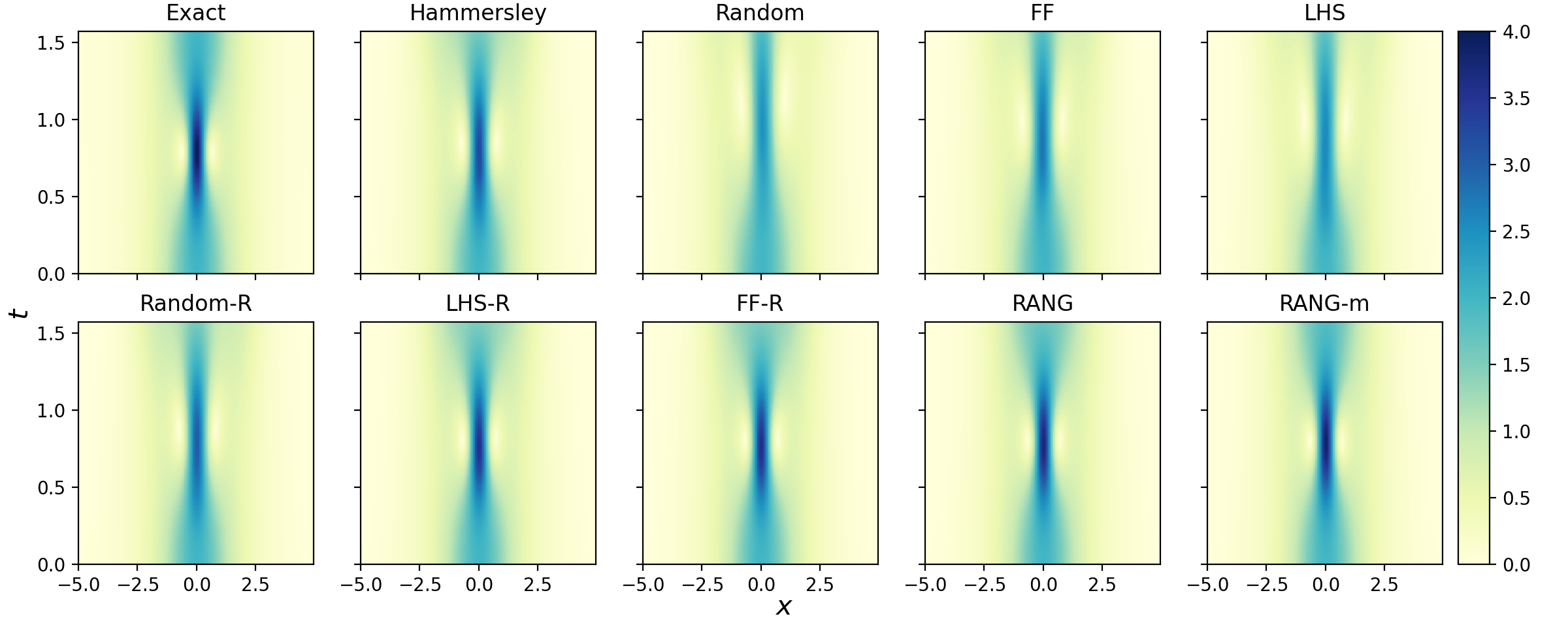}
		\caption{Solutions for the sch\"odinger equation. The top left image exhibits the reference solution. The remaining nine images show the prediction results of $9$ sampling methods in one experiment. The results might be different in another run due to the randomness of sampling and training.
		}
		\label{fig:essemble_schodinger}
	\end{figure}

\begin{figure}[htbp]
		\centering
		\includegraphics[width=1\textwidth]{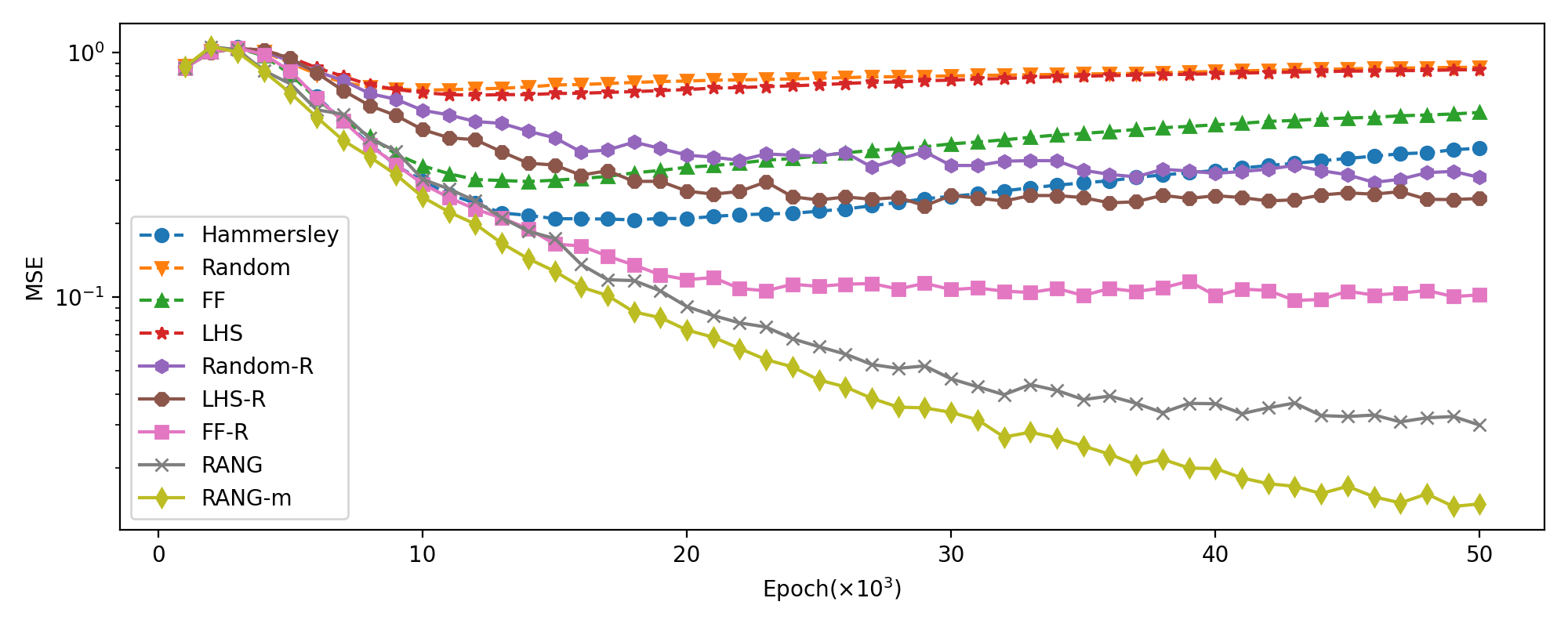}
		\caption{MSE values in iteration for the Sch\"odinger equation. This figure shows the curve of the median of MSE corresponding to nine different methods for the Sch\"odinger equation.}
		\label{fig:schodinger_it}
	\end{figure}

\begin{figure}[htbp]
		\centering
		\includegraphics[width=1\textwidth]{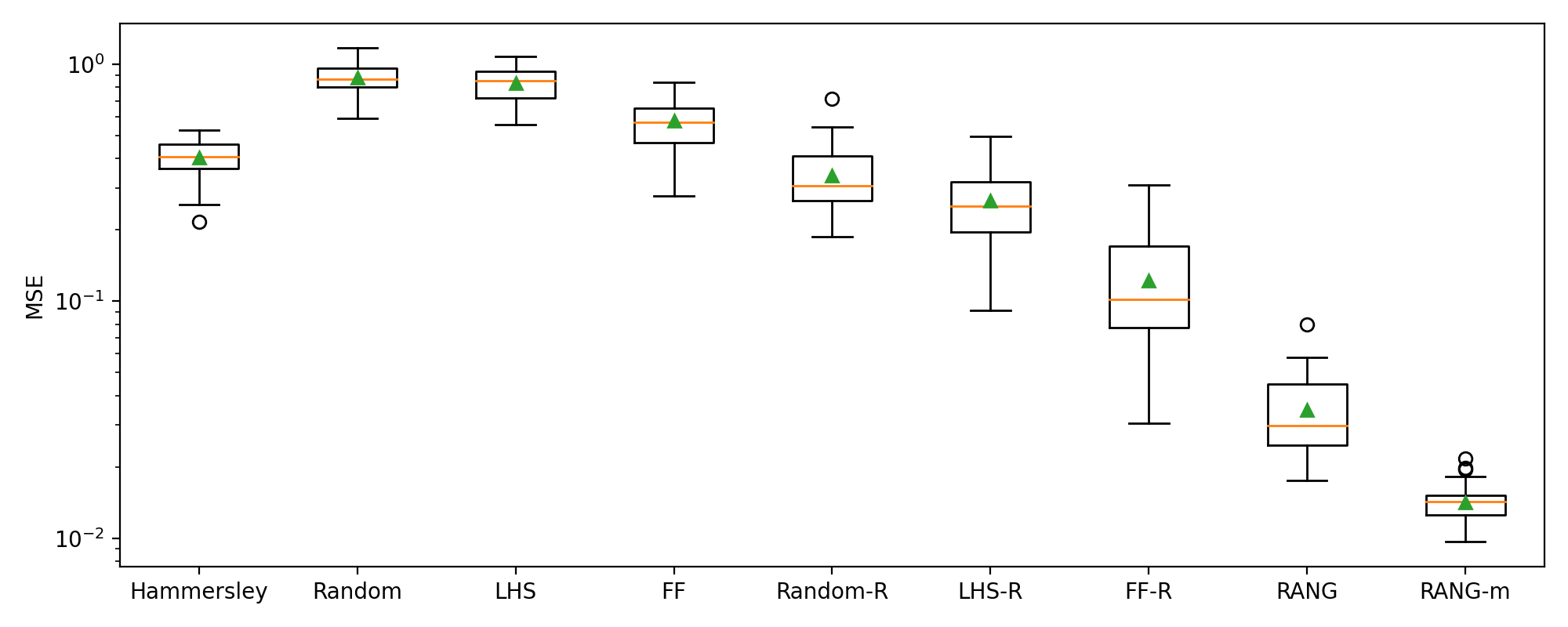}
		\caption{Box plot of MSEs for the Sch\"odinger equation considering different node generation strategies. The box plot represents median and 25, 75\% interquartile and extreme values of MSE for the sch\"odinger equation. The mean values are marked with triangles.}
		\label{fig:schodinger_error}
	\end{figure}

	The maximum number of iterations is set to $50,000$, and the number of replicates is $30$. The designated collocation point number $N_{pde}$ is set to $1,000$.
	Figure \ref{fig:essemble_schodinger} depicts the reference solution and results of the nine different sampling methods.
	As is shown in Figure \ref{fig:schodinger_it}, MSE curves of the four non-resampling methods decrease in the early iterations but increase later. The U-shape might be caused by the limited training collocation points, which leads to over-fitting after several iterations.
	The resampling methods seem to avoid this issue, but Random-R, FF-R, and LHS-R quickly fall into a plateau period, and the MSE no longer declines. 
	The two adaptive sampling methods, RANG and RANG-m, converge fast, and RANG-m endowed with the memory mechanism is better than RANG.
	To compare the methods, the MSE obtained by various sampling methods is shown in Figure \ref{fig:schodinger_error} and Table \ref{tab:schodinger} after training ends. All statistics indicate that RANG-m achieves the best performance. 
	
	\begin{table}
		\begin{tabular}{l|rrrrrr}
			\hline
			{} &      mean &       min &       25\% &       50\% &       75\% &       max \\
			\hline
			Hammersley & 4.038E-01 & 2.166E-01 & 3.639E-01 & 4.053E-01 & 4.591E-01 & 5.248E-01 \\
			Random     & 8.748E-01 & 5.907E-01 & 7.978E-01 & 8.672E-01 & 9.631E-01 & 1.167E+00 \\
			LHS        & 8.303E-01 & 5.554E-01 & 7.201E-01 & 8.488E-01 & 9.318E-01 & 1.080E+00 \\
			FF         & 5.760E-01 & 2.782E-01 & 4.678E-01 & 5.675E-01 & 6.523E-01 & 8.354E-01 \\
			Random-R   & 3.376E-01 & 1.864E-01 & 2.663E-01 & 3.074E-01 & 4.106E-01 & 7.143E-01 \\
			LHS-R      & 2.648E-01 & 9.134E-02 & 1.957E-01 & 2.525E-01 & 3.181E-01 & 4.969E-01 \\
			FF-R       & 1.225E-01 & 3.058E-02 & 7.762E-02 & 1.018E-01 & 1.713E-01 & 3.098E-01 \\
			RANG       & 3.475E-02 & 1.756E-02 & 2.473E-02 & 2.986E-02 & 4.477E-02 & 7.993E-02 \\
			RANG-m     & \textbf{1.415E-02} & \textbf{9.669E-03} & \textbf{1.256E-02} & \textbf{1.422E-02} & \textbf{1.520E-02} & \textbf{2.173E-02} \\
			\hline
		\end{tabular}
		\caption{The statistics of MSE for the Sch\"odinger equation considering different node generation strategies over 30 replicates. The optimal value is marked in bold for each statistic. The RANG-m method performs best in various statistics.}
		\label{tab:schodinger}
	\end{table}
	
	\subsection{Korteweg-de Vries Equation}
\begin{figure}[htbp]
		\centering
		\includegraphics[width=1\textwidth]{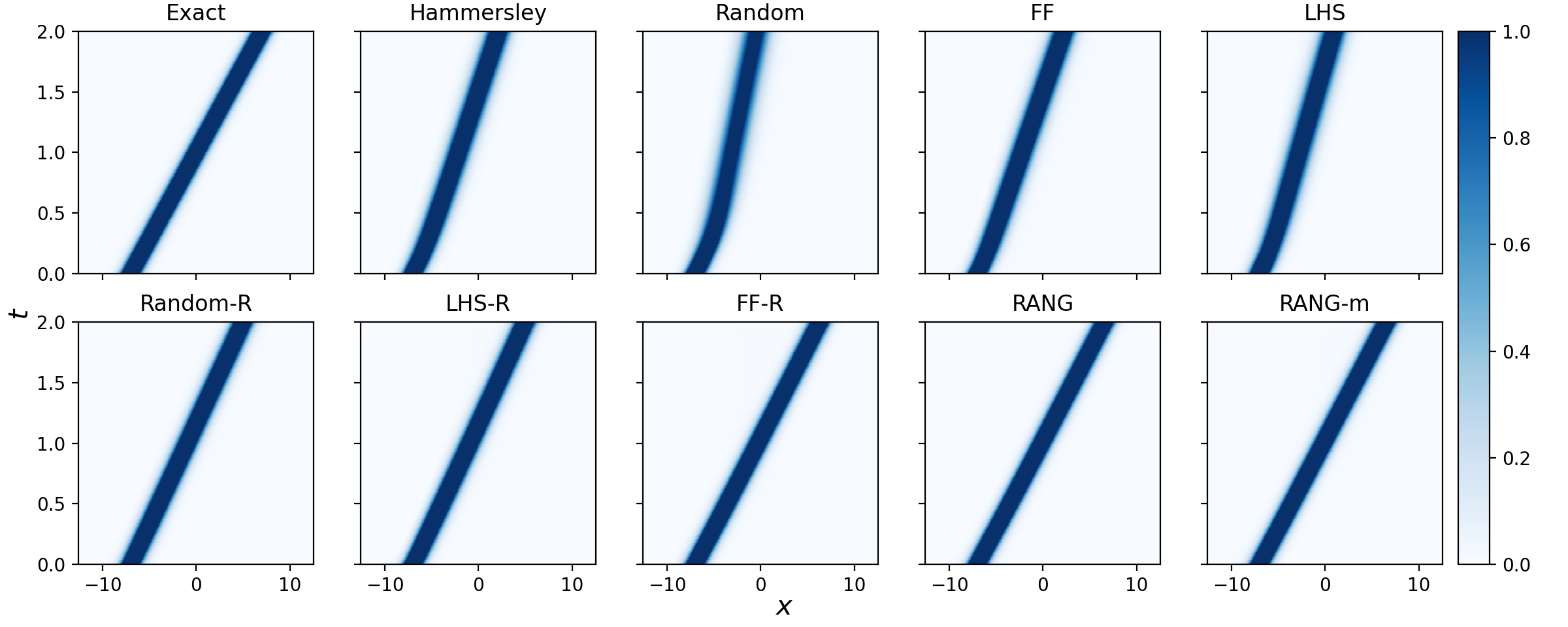}
		\caption{Solutions for the KdV equation. The top left image exhibits the reference solution of the KdV equation. The remaining nine images show the prediction results of nine sampling methods in one experiment. The results might be different in another run due to the randomness of sampling and training. The four non-resampling methods in the first row have significant errors.}
		\label{fig:essemble_kdv}
	\end{figure}

	The last nonlinear PDE we consider is the 1D Korteweg-de Vries (KdV) Equation.
	Its governing equation includes high-order derivatives. We consider the KdV equation in the following form:
	\begin{align*}
		\frac{\partial u}{\partial t} + 6u \frac{\partial u}{\partial x} + \frac{\partial^3u}{\partial x^3}=0,
	\end{align*}
	where the solitary wave solution is
	\begin{align*}
		u(t,x)=\frac{c}{2} \mathrm{sech}\left(\frac{\sqrt{c}}{2}(x - ct)\right)^2.
	\end{align*}
	Take $c=7.0$ and define the initial value condition:
	\begin{align*}
		u(0,x)=\frac{c}{2} \mathrm{sech}\left(\frac{\sqrt{c}}{2}(x+7)\right)^2.
	\end{align*}
	To handle the boundary value condition, we define a wide spatial domain $[-4\pi, 4\pi]$, where $u$ is nearly $0$ at two ends:
	\begin{align*}
		u(t,-4\pi)&=0,\\
		u(t,4\pi)&=0, \quad t\in[0,2].
	\end{align*}
	Define the initial and boundary value condition loss terms like the previous subsection, and we obtain the weighted total loss:
	\begin{align*}
		L=2L_0+2L_b+0.2L_{pde}.
	\end{align*}
	The maximum iteration number is $50,000$, and the number of replicates is $50$. The capacity of collocation point set is $N_{pde}=1000$.  Figure \ref{fig:essemble_kdv} depicts the reference solution and the results of nine different sampling methods after training ends.
	
	
\begin{figure}[htbp]
		\centering
		\includegraphics[width=1\textwidth]{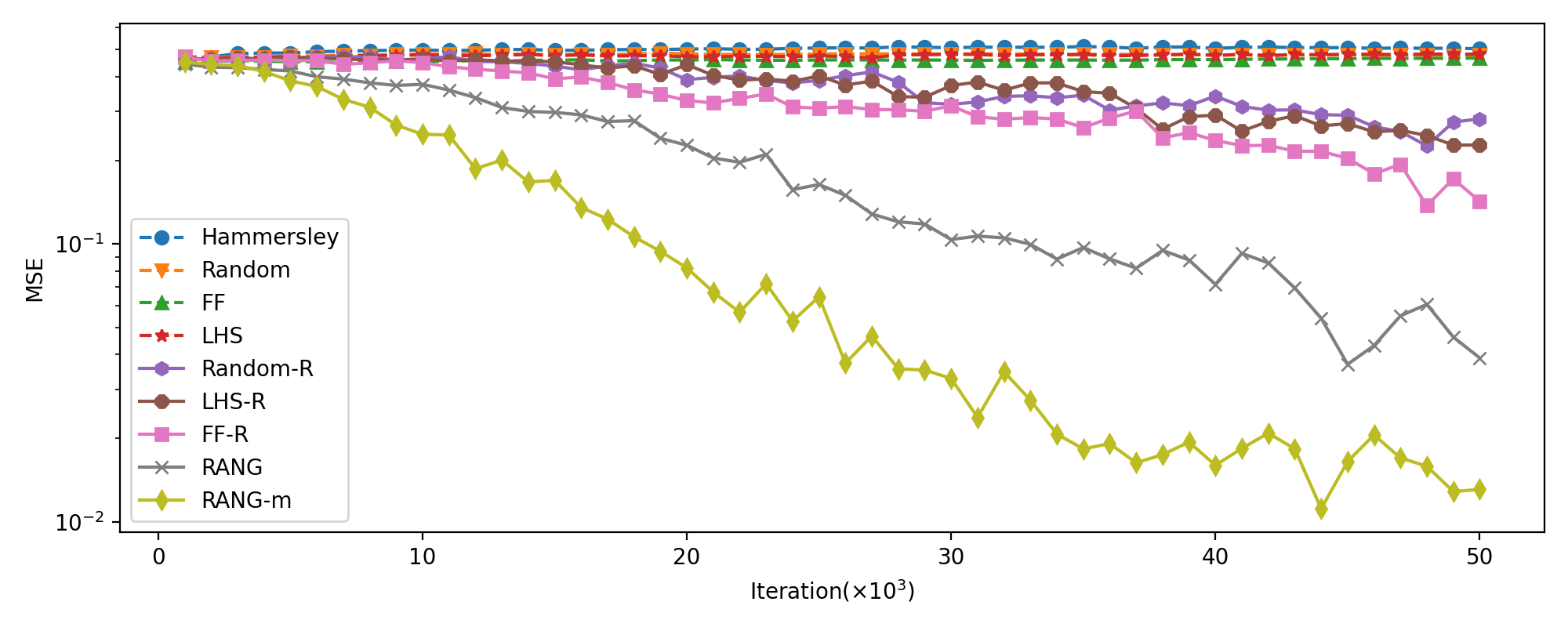}
		\caption{
			MSE values in iteration for the KdV equation. This figure shows the curve of the median of MSE corresponding to nine different methods for the KdV equation. Note that the four methods (Random/LHS/Random-R/LHS-R) are almost overlapped, and there is no decrease in their MSE values. MSE of the remaining methods decline, which might benefit from the local quasi-uniform properties.
		}
		\label{fig:kdv_it}
	\end{figure}

	As is shown in Figure \ref{fig:kdv_it}, it is hard for the four non-resampling methods to converge to the exact solution, which all fail to capture the motion of the solitary wave solution in Figure \ref{fig:essemble_kdv}. The two adaptive methods, RANG and RANG-m, perform best among the nine methods.
	
\begin{figure}[htbp]
		\centering
		\includegraphics[width=1\textwidth]{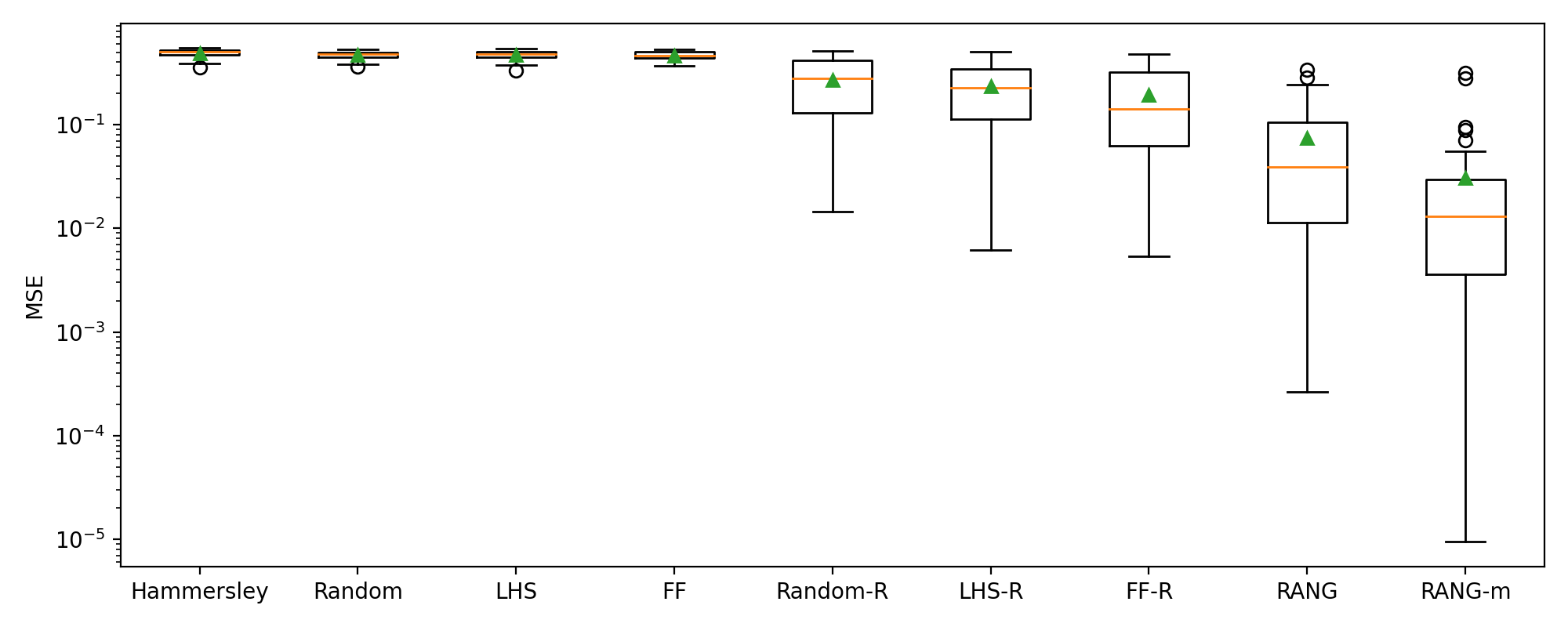}
		\caption{Box plot of MSEs for the KdV equation considering different node generation strategies. The plot represents median and 25, 75\% interquartile and extreme values. The mean values are marked with triangles.}
		\label{fig:kdv_error}
	\end{figure}

	MSE values obtained by nine sampling methods are shown in Figure \ref{fig:kdv_error} and Table \ref{tab:kdv}. All statistics indicate that RANG-m achieves the best performance. 
	\begin{table}
		\begin{tabular}{l|rrrrrr}
			\hline
			{} &      mean &       min &       25\% &       50\% &       75\% &       max \\
			\hline
			Hammersley & 4.913E-01 & 3.567E-01 & 4.679E-01 & 5.043E-01 & 5.280E-01 & 5.471E-01 \\
			Random     & 4.699E-01 & 3.617E-01 & 4.486E-01 & 4.775E-01 & 4.976E-01 & 5.313E-01 \\
			LHS        & 4.712E-01 & 3.316E-01 & 4.457E-01 & 4.816E-01 & 5.073E-01 & 5.408E-01 \\
			FF         & 4.664E-01 & 3.690E-01 & 4.410E-01 & 4.665E-01 & 5.026E-01 & 5.342E-01 \\
			Random-R   & 2.679E-01 & 1.444E-02 & 1.293E-01 & 2.820E-01 & 4.155E-01 & 5.136E-01 \\
			LHS-R      & 2.346E-01 & 6.175E-03 & 1.138E-01 & 2.268E-01 & 3.423E-01 & 5.039E-01 \\
			FF-R       & 1.926E-01 & 5.381E-03 & 6.253E-02 & 1.422E-01 & 3.206E-01 & 4.792E-01 \\
			RANG       & 7.387E-02 & 2.642E-04 & 1.137E-02 & 3.888E-02 & 1.049E-01 & 3.380E-01 \\
			RANG-m     & \textbf{3.050E-02} & \textbf{9.484E-06} & \textbf{3.574E-03} & \textbf{1.309E-02} & \textbf{2.979E-02} & \textbf{3.171E-01} \\
			\hline
		\end{tabular}
		\caption{The statistics of MSE for the KdV equation considering different node generation strategies over $50$ replicates. The optimal value is marked in bold for each statistic. The RANG-m method performs best in various statistics.}
		\label{tab:kdv}
	\end{table}

	\subsection{2D Poisson Equation}
	The second linear PDE we consider is a 2D Poisson equation in a square:
	\begin{align*}
		\frac{\partial^2 u}{\partial x^2}+\frac{\partial^2 u}{\partial y^2}&=w_{\sigma}(x,y),\quad (x,y)\in\Omega=[-1,1]\times[-1,1]\\
		u|_{\partial \Omega}&=0,
	\end{align*}
	where
	\begin{align*}
		w_\sigma(x,y)=\frac{-2\sigma^2 + (x-0.3)^2+(y-0.3)^2}{\sigma^4}\exp\left(-\frac{(x-0.3)^2 + (y-0.3)^2}{2\sigma^2}\right)\\-\frac{-2\sigma^2 + (x+0.3)^2+(y+0.3)^2}{\sigma^4}\exp\left(-\frac{(x+0.3)^2 + (y+0.3)^2}{2\sigma^2}\right).
	\end{align*}
	The ground truth is the superposition of two opposite Gaussian pulses:
	\begin{align*}
		u(x,y) = g_\sigma(x-0.3,y-0.3)- g_\sigma(x+0.3,y+0.3)
	\end{align*}
	where the Gaussian pulse function $g_\sigma(x,y)$ is defined by
	\begin{align*}
		g_\sigma(x,y)=\exp\left(-\frac{x^2 + y ^2}{2\sigma^2}\right).
	\end{align*}
	Due to the lack of the initial value conditions, only the boundary condition loss $L_b$ and the PDE loss $L_{pde}$ are taken into consideration. Take $\sigma=0.1$, and the solution $u$ restricted on the boundary is nearly zero. Then we define the boundary value loss term:
	\begin{align*}
		L_b = \frac{1}{4\#\mathbf{N}_b}\sum_{x_b\in\mathbf{N}_b}\left(|u(x_b,-1)|^2+|u(x_b,1)|^2+|u(-1,x_b)|^2+|u(1,x_b)|^2\right).
	\end{align*}
	The PDE loss term is defined as 
	\begin{align*}
		L_{pde} = \frac{1}{\#\mathbf{M}_j}\sum_{(\tilde x,\tilde y)\in \mathbf{M}_{j}}\left|\frac{\partial^2 u}{\partial x^2}(\tilde x,\tilde y)+\frac{\partial^2 u}{\partial y^2}(\tilde x,\tilde y)-g_\sigma(\tilde x,\tilde y)\right|^2.
	\end{align*}
	The weighted total loss is
	\begin{align*}
		L=2\cdot L_b+0.2\cdot L_{pde}.
	\end{align*}
	The neuron number for each layer is $[2,64,64,64,64,1]$.
	The maximum iteration number is set to 3,000, and the resampling interval $I$ is set to $100$. The number of replicates is $100$.
	The designated capacity of collocation point set is $N_{pde}=400$. 
	
\begin{figure}[htbp]
		\centering
		\includegraphics[width=1\textwidth]{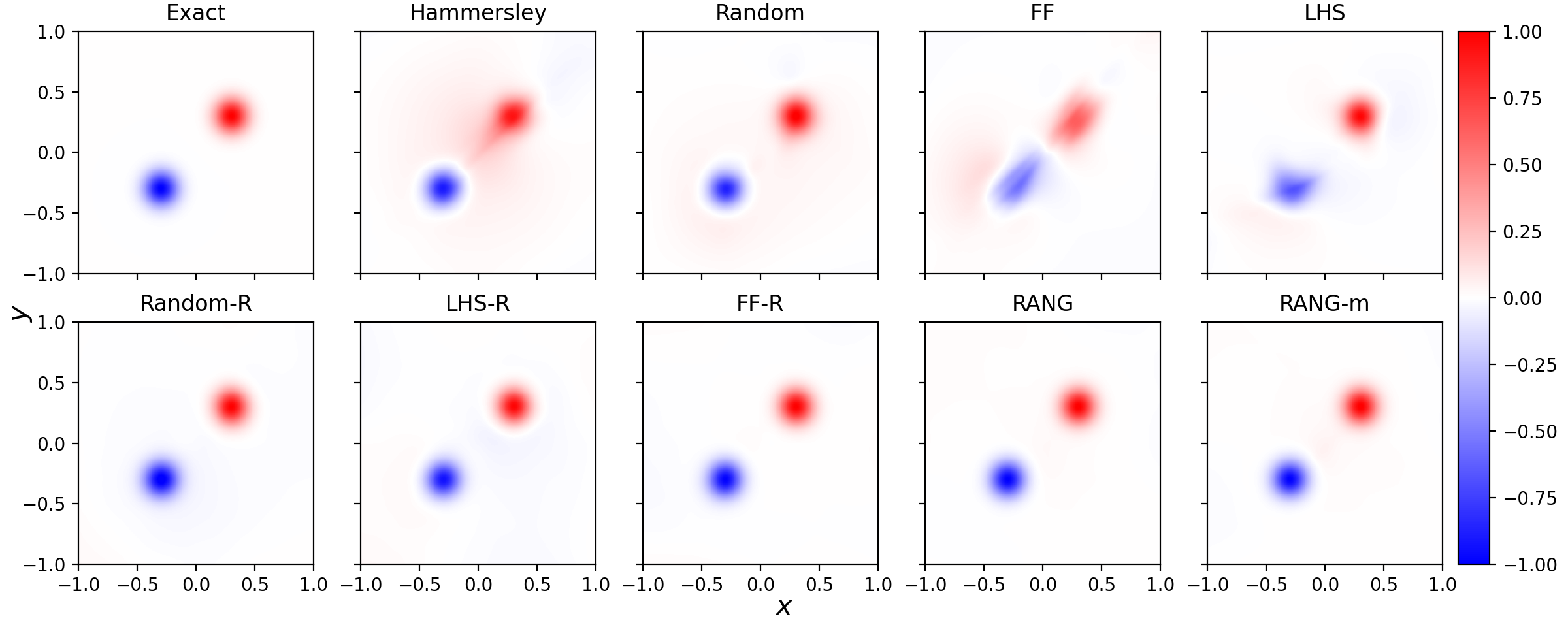}
		\caption{Solutions for the Poisson equation. The top left image exhibits the reference solution of the Poisson equation. The remaining nine images show the prediction results of nine sampling methods in one experiment. The results might be different in another run due to the randomness of sampling and training.
		}
		\label{fig:essemble_poisson}
	\end{figure}

	Figure \ref{fig:essemble_poisson} depicts the reference solution and the results of nine different sampling methods.
	
\begin{figure}[htbp]
		\centering
		\includegraphics[width=1\textwidth]{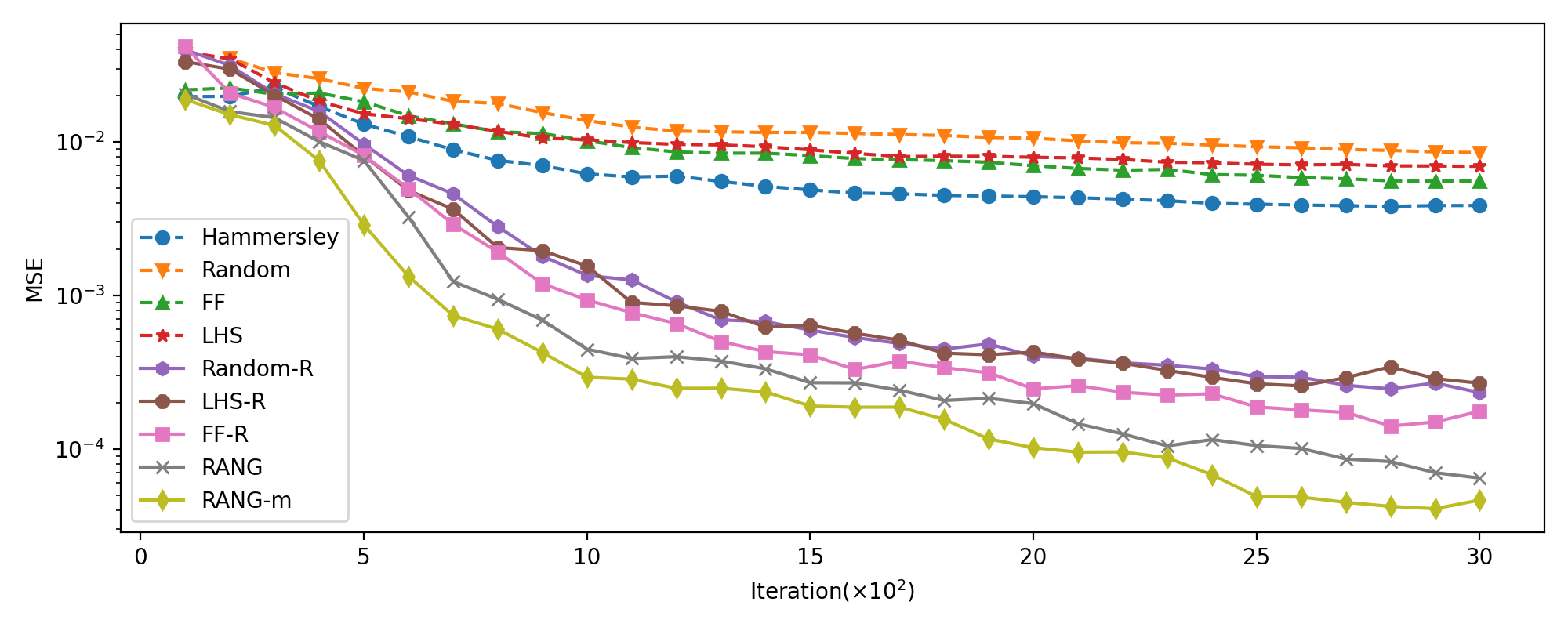}
		\caption{MSE values in iteration for the Poisson equation. This figure shows the curve of the median of MSE corresponding to the nine different methods for the Poisson equation. The four non-resampling methods (Hammersley/Random/FF/LHS) are obviously slower than the five resampling methods.}
		\label{fig:poisson_it}
	\end{figure}

\begin{figure}[htbp]
		\centering
		\includegraphics[width=1\textwidth]{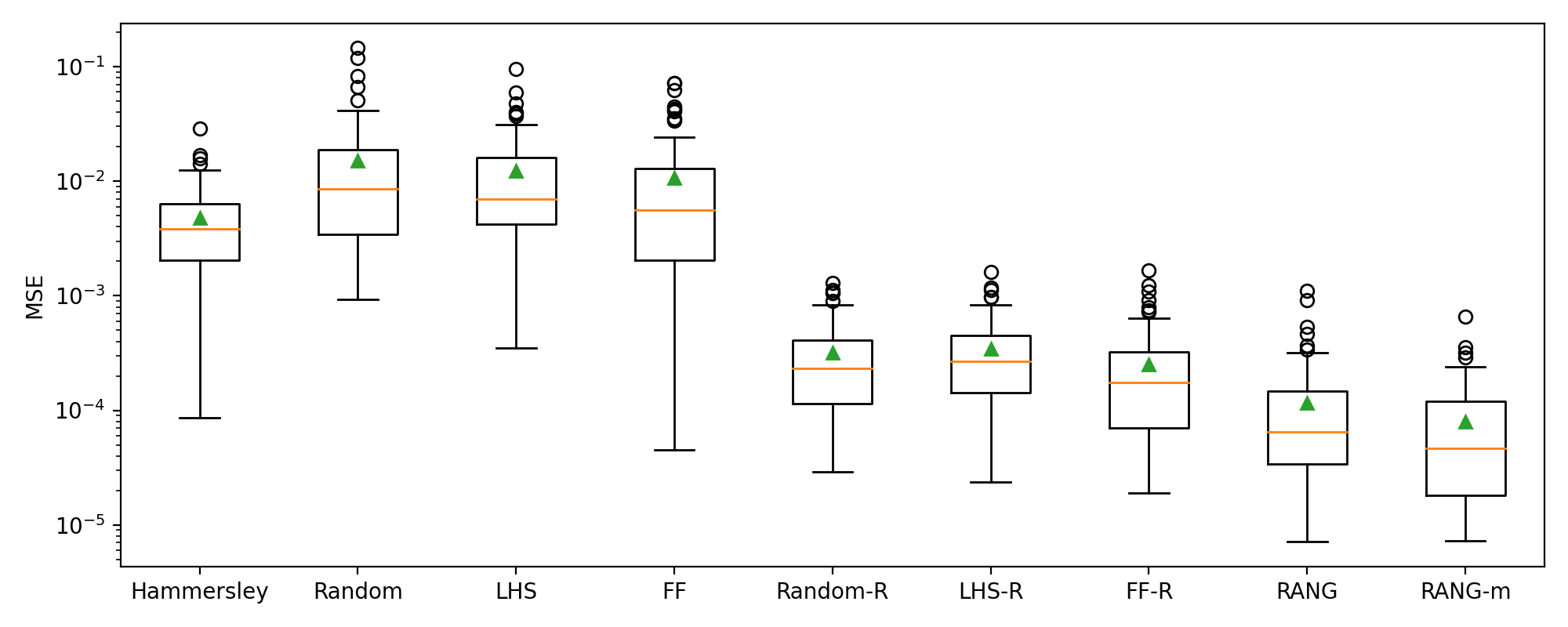}
		\caption{Box plot of MSEs for the Poisson equation considering different node generation strategies. The box represents median and 25, 75\% interquartile and extreme values. The mean values are marked with triangles.}
		\label{fig:poisson_error}
	\end{figure}

	As shown in Figure \ref{fig:poisson_it}, the error curves of the four non-resampling methods decline slowly. RANG and RANG-m perform better than the others. 
	After 3,000 iteration, the MSE obtained by various sampling methods is shown in Figure \ref{fig:poisson_error} and Table \ref{tab:poisson}.

	\begin{table}
		\begin{tabular}{lrrrrrr}
			\hline
			{} &      mean &       min &       25\% &       50\% &       75\% &       max \\
			\hline
			Hammersley & 4.805E-03 & 8.556E-05 & 2.034E-03 & 3.850E-03 & 6.379E-03 & 2.874E-02 \\
			Random     & 1.517E-02 & 9.231E-04 & 3.428E-03 & 8.514E-03 & 1.868E-02 & 1.448E-01 \\
			LHS        & 1.225E-02 & 3.490E-04 & 4.204E-03 & 6.928E-03 & 1.606E-02 & 9.605E-02 \\
			FF         & 1.076E-02 & 4.498E-05 & 2.041E-03 & 5.562E-03 & 1.280E-02 & 7.144E-02 \\
			Random-R   & 3.185E-04 & 2.899E-05 & 1.137E-04 & 2.325E-04 & 4.086E-04 & 1.302E-03 \\
			LHS-R      & 3.420E-04 & 2.383E-05 & 1.417E-04 & 2.679E-04 & 4.482E-04 & 1.624E-03 \\
			FF-R       & 2.528E-04 & 1.889E-05 & 6.966E-05 & 1.760E-04 & 3.225E-04 & 1.667E-03 \\
			RANG       & 1.158E-04 & \textbf{7.146E-06} & 3.417E-05 & 6.461E-05 & 1.474E-04 & 1.102E-03 \\
			RANG-m     & \textbf{7.970E-05} & 7.286E-06 & \textbf{1.818E-05} & \textbf{4.645E-05} & \textbf{1.192E-04} & \textbf{6.610E-04} \\
			\hline
		\end{tabular}
		\caption{The statistics of MSE for the Poisson equation considering different node generation strategies over $100$ replicates. The optimal value is marked in bold for each statistic. The RANG-m method performs best in various statistics except for the extreme minimum, which is slightly larger than RANG.}
		\label{tab:poisson}
	\end{table}
	
	\subsection{1D convection–diffusion equation Equation}
	The last linear PDE in the section is a 1D convection-diffusion equations containing an advection term and a diffusion term. The governing PDE is shown as follows:
	\begin{align*}
		\frac{\partial u}{\partial t}+c\cdot\frac{\partial u}{\partial x}&=\mu\cdot\frac{\partial^2 u}{\partial x^2},\\
		u(0,x)&=u_0(x),\\
		(t,x)&\in[0,T]\times(-\infty, \infty).
	\end{align*}
	Take $T=1.0$, $c=4.0$ and $\mu=0.05$, the initial value is a Gaussian pulse function
	\begin{align*}
		u_0(x)=\frac{0.1}{\sqrt{0.1\mu}}\cdot\exp\left(-\frac{(x+2)^2}{4\cdot  0.1\cdot \mu}\right).
	\end{align*}
	And the exact solution is
	\begin{align*}
		u_0(x)=\frac{0.1}{\sqrt{\mu(t + 0.1)}} \cdot \exp\left(-\frac{(x+2-4t)^2}{4 \mu (t + 0.1))}\right).
	\end{align*}
	We truncate the spatial domain to $[-l, l]$ such that the solution is almost $0$ at the two ends. The boundary value condition is defined as
	\begin{align*}
		u(t,-l)=u(t,l)=0, t\in[0,T].
	\end{align*}
	Then we solve the equation in domain $[0,T]\times[-l,l]$.
	Following the steps of previous subsections, we get the weighted total loss:
	\begin{align*}
		L=2L_0+2L_b+0.2L_{pde}.
	\end{align*}
	The neuron number for each layer is $[2,64,64,64,64,1]$.
	The maximum iteration number is set to 10,000, and the resample interval $I$ is set to 1,000. The number of replicates is $60$. The designated capacity of collocation point set is $N_{pde}=1000$.

	\begin{figure}[htbp]
		\centering
		\includegraphics[width=1\textwidth]{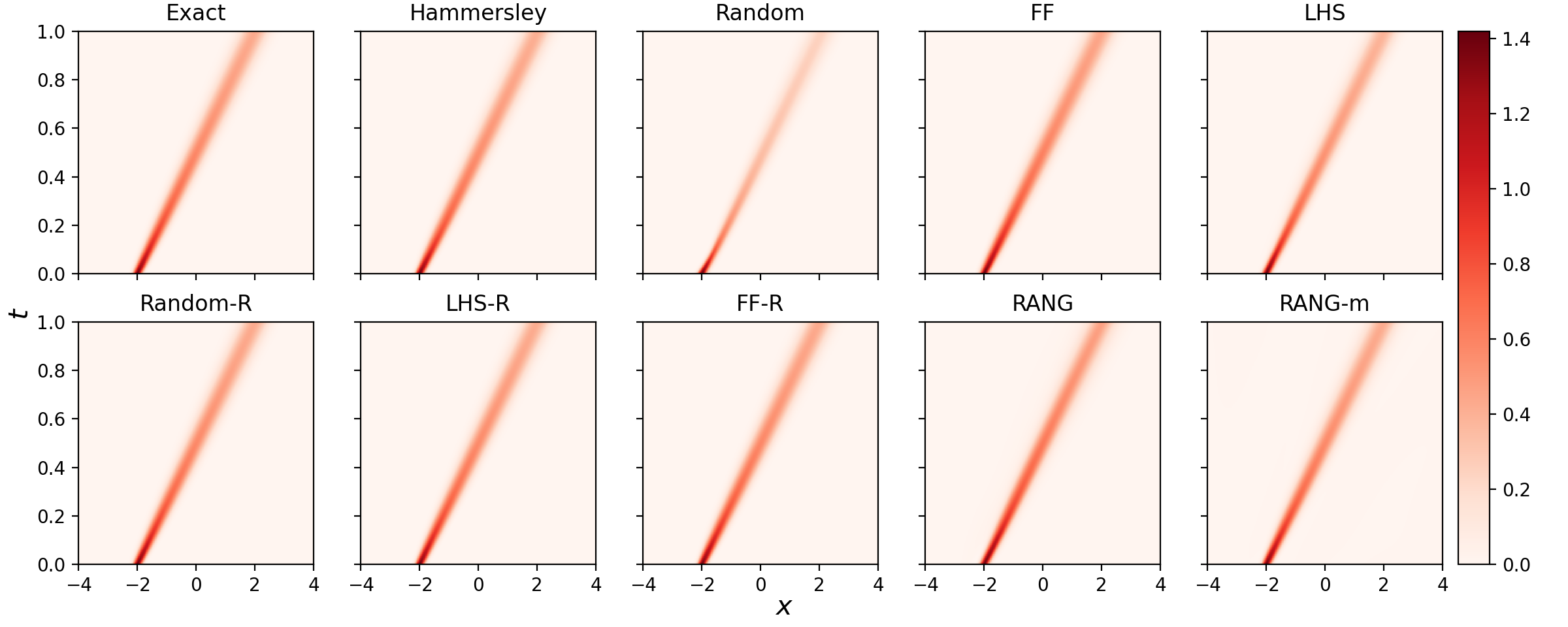}
		\caption{Solutions for the convection–diffusion equation. The top left image exhibits the reference solution. The remaining nine images show the prediction results of nine sampling methods in one experiment. The results might be different in another run due to the randomness of sampling and training.}
		\label{fig:essemble_heat}
	\end{figure}
\begin{figure}[htbp]
		\centering
		\includegraphics[width=1\textwidth]{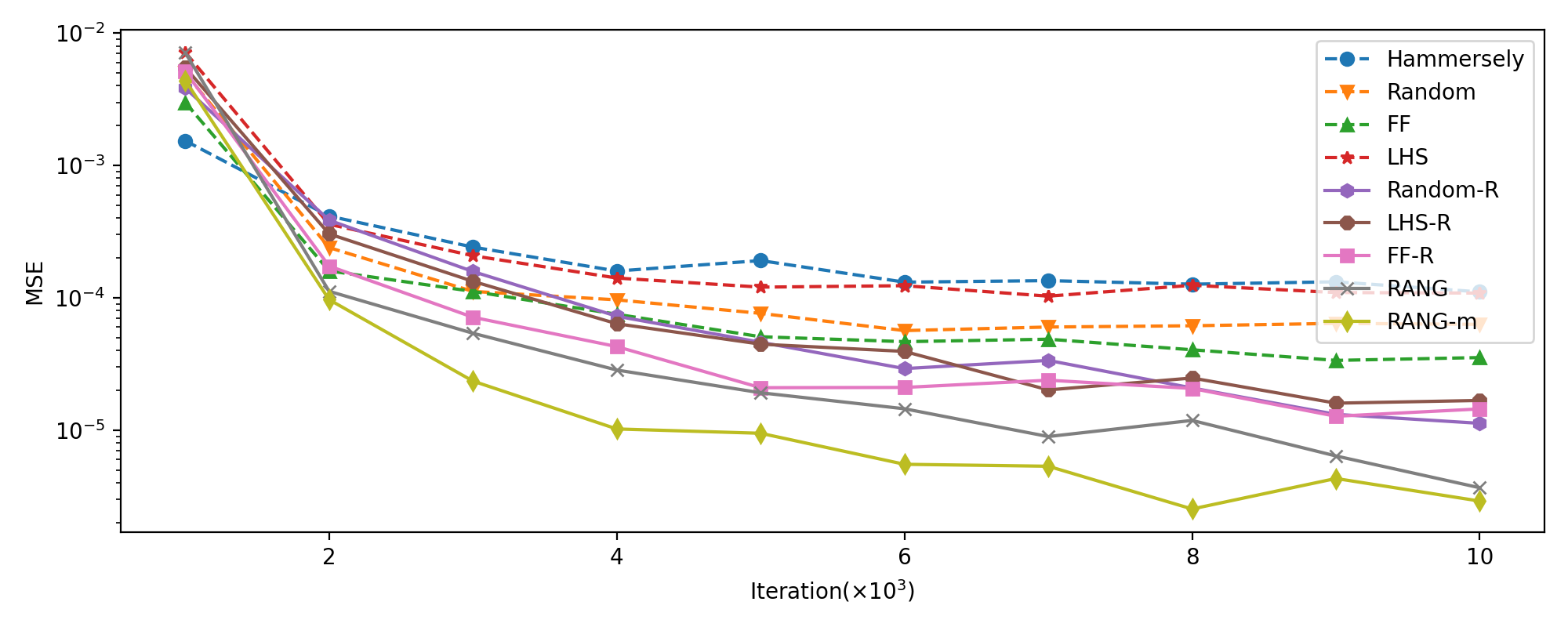}
		\caption{MSE values in iteration for the convection–diffusion equation. This figure shows the curve of the median of MSE corresponding to nine different methods.}
		\label{fig:advection_it}
	\end{figure}
\begin{figure}[htbp]
		\centering
		\includegraphics[width=1\textwidth]{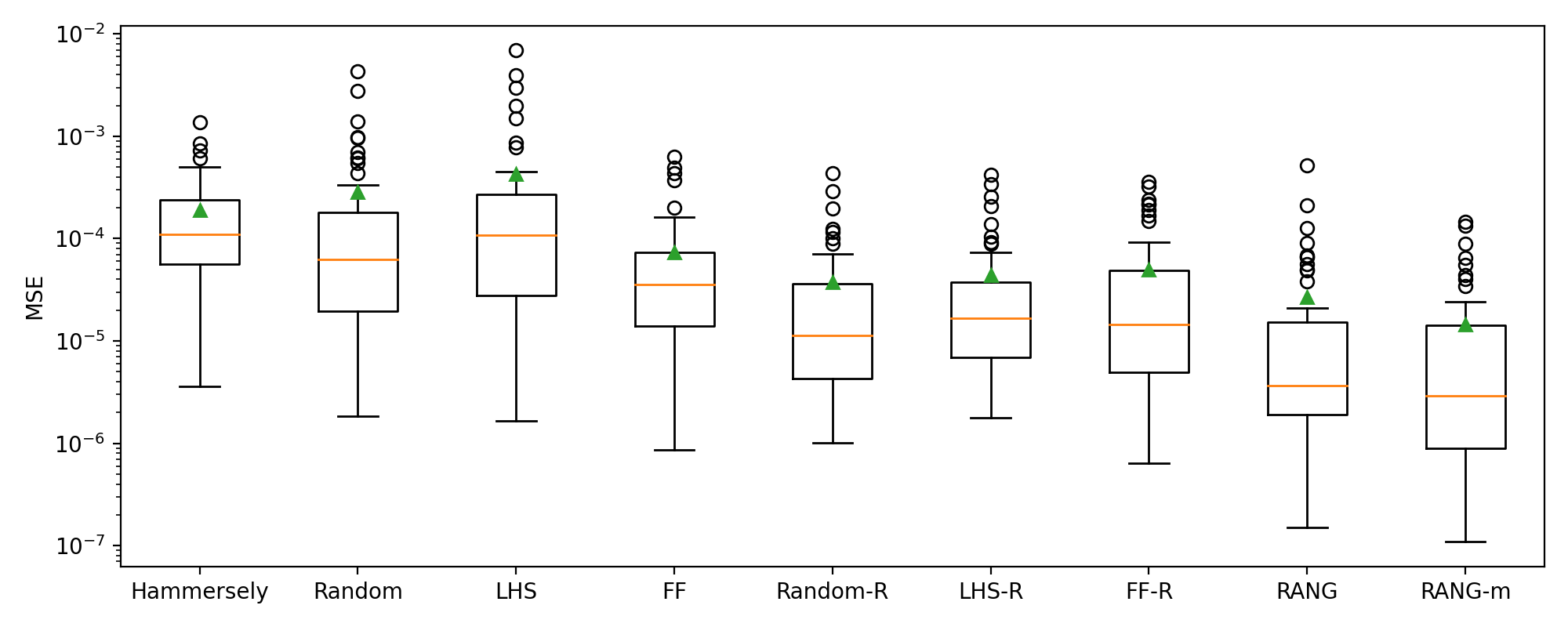}
		\caption{Box plot of MSEs for the convection–diffusion equation considering different node generation strategies. The plot represents median and 25, 75\% interquartile and extreme values. The mean values are marked with triangles.}
		\label{fig:advection_error}
	\end{figure}

	As shown in Figure \ref{fig:advection_it}, the error curves of the four non-resampling methods decline slowly. Three non-adaptive resampling methods are faster. RANG and RANG-m are still the fastest.
	After training ends, the MSE obtained by various sampling methods is shown in Figure \ref{fig:advection_error} and Table \ref{tab:adv}. All statistics imply that RANG-m achieves the best performance. 
	\begin{table}
		\begin{tabular}{l|rrrrrr}
			\hline
			{} &      mean &           min &           25\% &       50\% &       75\% &       max \\
			\hline
			Hammersley & 1.916E-04 & 3.582E-06 & 5.633E-05 & 1.109E-04 & 2.392E-04 & 1.368E-03 \\
			Random     & 2.842E-04 & 1.839E-06 & 1.965E-05 & 6.302E-05 & 1.817E-04 & 4.317E-03 \\
			LHS        & 4.286E-04 & 1.654E-06 & 2.785E-05 & 1.085E-04 & 2.728E-04 & 6.958E-03 \\
			FF         & 7.311E-05 & 8.566E-07 & 1.408E-05 & 3.535E-05 & 7.370E-05 & 6.284E-04 \\
			Random-R   & 3.789E-05 & 1.007E-06 & 4.266E-06 & 1.126E-05 & 3.638E-05 & 4.377E-04 \\
			LHS-R      & 4.412E-05 & 1.773E-06 & 6.903E-06 & 1.677E-05 & 3.779E-05 & 4.216E-04 \\
			FF-R       & 5.003E-05 & 6.388E-07 & 4.961E-06 & 1.447E-05 & 4.875E-05 & 3.592E-04 \\
			RANG       & 2.671E-05 & 1.510E-07 & 1.899E-06 & 3.671E-06 & 1.528E-05 & 5.183E-04 \\
			RANG-m     & \textbf{1.459E-05} & \textbf{1.093E-07} & \textbf{8.959E-07} & \textbf{2.912E-06} & \textbf{1.422E-05} & \textbf{1.453E-04} \\
			\hline
		\end{tabular}
		\caption{
			The statistics of MSE for the convection–diffusion equation considering different node generation strategies over $60$ replicates. The optimal value is marked in bold for each statistic. The RANG-m method performs best in various statistics, close to RANG.}
		\label{tab:adv}
	\end{table}

	\section{Conclusive Remarks}
	
	We propose a residual-based adaptive node generation method for training PINN on a two-dimensional domain, enjoying local regularity and adaptivity simultaneously. The memory mechanism is included to enhance the training stability. 
	A variety of sampling methods on linear and nonlinear PDEs are taken into comparison. The results show that PINN with the proposed method can improve convergence.
	
	We observed that configurations of collocation nodes significantly impact the results, just like meshless methods such as RBF-FD. PINN has similar or different requirements from RBF-FD for evaluating the quality of collocation points. 
	Several aspects are compared:
	\begin{enumerate}
		\item \textbf{Minimum distance}. Neural networks usually have a large number of parameters and are iteratively trained by gradient descent with small learning rate. It does not affect the stability of PINN even in the case of two training points coinciding. However, collocation nodes that are positioned too closely can severely impact the stability of some meshfree methods \cite{liuMeshFreeMethods2002}. Therefore, we can choose FF as the basic framework, even though it does not guarantee the minimum distance.
		\item \textbf{Local regularity}. For RBF-FD, nodal distributions should be locally regular throughout the computational domain, i.e., the distances between neighbor nodes should be roughly equal. The requirement derives from the fact that local strong form meshless methods are often sensitive to node positions, and large differences in distances to the nearest neighbors or other irregularities can cause ill-conditioned approximations, making the distribution improper for solving PDEs \cite{slakGenerationNodeDistributions2019a}. 
		Though the local regularity does not affect the stability of the minimization problem introduced by PINN, it is usually more efficient with the similar numbers of collocation nodes in our experiments. Random and LHS methods do not generate nodes with local regularity, while Hammersley and FF can generate locally regular nodes, which might be the reason why Hammersley and FF usually have better results than the other non-resampling methods in our experiments. 
		\item \textbf{Adaptive Node Generation}. Similar adaptive node generation methods have been proposed in the RBF-FD studies \cite{slakAdaptiveRadialBasis2019,liAdaptiveRBFFDMethod2017}, which illustrate that PINN may benefit from traditional meshless studies and vice versa.
	\end{enumerate}
	
	RBF-FD is a more mature meshless method, and many techniques used in RBF-FD might be transferred and applied to PINN studies. This method provides a preliminary attempt at node generation inspired by traditional meshless methods for PINNs.  
	
	Limited to the scope, this paper is aimed at a two-dimensional rectangular domain, which corresponds to a steady two-dimensional PDE on a rectangular space region or a time-dependent one-dimensional PDE. For high-dimensional or more complex regions, the following extensions should be considered:
	\begin{enumerate}
		\item For a two-dimensional complex domain, consider generating nodes in a large enough rectangular region containing the domain and filtering out the points outside, as is shown in Figure \ref{Lshape}, sampling on the square region, and removing the points outside the $L$-shape. Note that the method might be highly inefficient.
		\item Using subsequent methods of FF, the collocation points can be generated based on any given boundary \cite{slakGenerationNodeDistributions2019a}. For high-dimensional PDEs or PDEs on manifolds, the following methods of FF can replace the basis (FF) in RANG for implementation.
		\item Combined with the domain decomposition methods in PINN, complex domains are converted into several simple subdomains and then solved \cite{mengPPINNPararealPhysicsinformed2020,shuklaParallelPhysicsinformedNeural2021}. 
		The nodes can also be generated for each simple subdomain seperately. Since points that are very close do not undermine the stability of PINNs, and then no minimum distance guarantees are needed. Therefore, nodes generated near the interfaces and boundaries do not need the repel algorithm for post-processing.
		\item Adaptive sampling methods for boundary and initial condition residuals are not considered. The boundaries used in the experiments are all rectangular. Furthermore, in order to reduce the influence of random factors, the nodes for boundary or initial value loss terms are equispaced artificially in the experiments. However, for high-dimensional problems with complex regions, adaptively learning boundary conditions may also benefit from the proposed point generation methods combined with variable density manifold sampling methods \cite{duhFastVariableDensity2021}.
	\end{enumerate}
	
	\section*{Acknowledgement}
	This work was supported by National Natural Science Foundation of China under Grant No.52005505 and 11725211.
	
	\bibliographystyle{elsarticle-num} 
	\bibliography{cas-refs}
	
	
	
	
	
\end{document}